\documentclass{egpubl}
\usepackage{eg2021}
\usepackage{balance}

\ConferencePaper 
\usepackage[T1]{fontenc}
\usepackage{dfadobe}  
\usepackage{amsmath,amssymb}
\usepackage{color}

\usepackage{wrapfig}
\usepackage{makecell}
\usepackage{comment}
\usepackage{mathtools}
\usepackage{kotex}

\usepackage{float}
\usepackage{array,multirow,multicol}

\usepackage{cite} 
\BibtexOrBiblatex
\electronicVersion
\PrintedOrElectronic
\ifpdf \usepackage[pdftex]{graphicx} \pdfcompresslevel=9
\else \usepackage[dvips]{graphicx} \fi

\usepackage{egweblnk} 
\newcommand{\Eq}[1]  {Eq.\ (\ref{equ:#1})}

\newcommand{\Fig}[1] {Figure \ref{fig:#1}}
\newcommand{\Figs}[1]{Figures \ref{fig:#1}}
\newcommand{\Tbl}[1]  {Table \ref{tbl:#1}}

\newcommand{\Sec}[1] {Section \ref{sec:#1}}

\newcommand{\Etal}   {et al.}

\newcommand{\specialcell}[2][c]{%
  \begin{tabular}[#1]{@{}c@{}}#2\end{tabular}}

\title[Spatiotemporal Texture Reconstruction for Dynamic Objects Using a Single RGB-D Camera]%
      {Spatiotemporal Texture Reconstruction for Dynamic Objects \\
       Using a Single RGB-D Camera}
\author[Kim et al.]
{\parbox{\textwidth}{\centering Hyomin Kim\orcid{0000-0002-2162-4627}, Jungeon Kim\orcid{0000-0003-4212-1970}, Hyeonseo Nam\orcid{0000-0003-4033-901X}, Jaesik Park\orcid{0000-0001-5541-409X}, and Seungyong Lee\orcid{0000-0002-8159-4271}
	}
	\\
	{\parbox{\textwidth}{\centering POSTECH
		}
	}
}

\begin{document}

\teaser{
 \includegraphics[width=\linewidth]{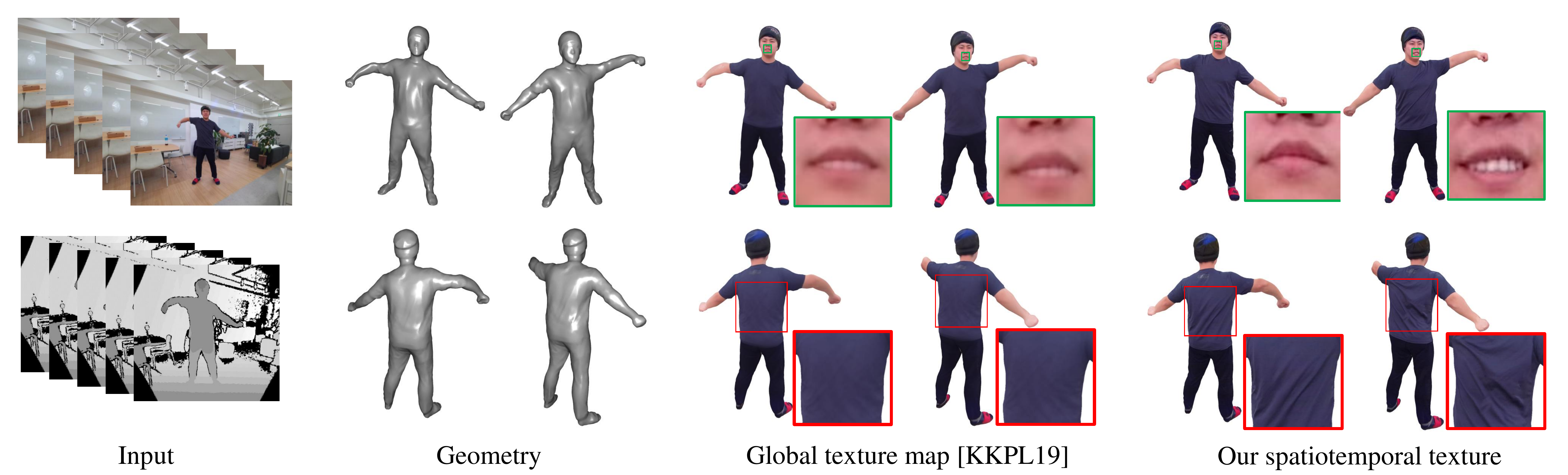}
 \centering
  \caption{Our approach reconstructs a time-varying (spatiotemporal) texture map for a dynamic object using partial observations obtained by a single RGB-D camera. The frontal and rear views (top and bottom rows) of the geometry at two frames are shown in the middle left. Compared to the global texture atlas-based approach~\cite{kim2019global}, our method produces more appealing appearance changes of the object. Please see the supplementary video for better visualization of time-varying textures.}
\label{fig:teaser}
}

\maketitle
\begin{abstract}
This paper presents an effective method for generating a spatiotemporal (time-varying) texture map for a dynamic object using a single RGB-D camera. The input of our framework is a 3D template model and an RGB-D image sequence. Since there are invisible areas of the object at a frame in a single-camera setup, textures of such areas need to be borrowed from other frames. We formulate the problem as an MRF optimization and define cost functions to reconstruct a plausible spatiotemporal texture for a dynamic object. Experimental results demonstrate that our spatiotemporal textures can reproduce the active appearances of captured objects better than approaches using a single texture map.


\end{abstract}  
\section{Introduction}
\label{sec:introduction}

3D reconstruction methods using RGB-D images have been developed for static scenes~\cite{newcombe2011kinectfusion, voxelHashing, dai2017bundlefusion} and dynamic objects~\cite{newcombe2015dynamicfusion, innmann2016volumedeform, fusion4d}.
The appearance of the reconstructed 3D models, which are often represented as color meshes, is crucial in realistic content creation for Augmented Reality (AR) and Virtual Reality (VR) applications. Subsequent works have been proposed to improve the color quality of reconstructed static scenes~\cite{Seamless_Montage, ZhouColoMap, jeon2016texture,fu2018texture,Li2019FastTexMapping_tvcg} and dynamic objects~\cite{orts2016holoportation, du2018montage4d, guo2019relightables}.
For static scenes, most of the approaches generate a global texture atlas~\cite{waechter2014let, jeon2016texture, fu2018texture, Li2019FastTexMapping_tvcg}, and
the main tasks consist of two parts: selecting the best texture patches from input color images and relieving visual artifacts such as texture misalignments and illumination differences when the selected texture patches are combined.

While single texture map is widely used to represent a static scene, it is not preferred for a dynamic object due to appearance changes over time. A multi-camera system could facilitate creating a spatiotemporal (time-varying) texture map of a dynamic object, but such setup may not be affordable for ordinary users. As an alternative, recovering a spatiotemporal texture using a \emph{single RGB-D camera} can be a practical solution.

Previous reconstruction methods using a single RGB-D camera maintain 
and update voxel colors in the canonical frame~\cite{newcombe2015dynamicfusion, innmann2016volumedeform,Guo2017realGeoAlbe, BodyFusion, DoubleFusion}. However, this voxel color based approach cannot properly represent detailed time-varying appearances, e.g., cloth wrinkles and facial expression changes for capturing humans.
Kim~\Etal~\cite{kim2019global} proposed a multi-scale optimization method to reconstruct a global texture map of a dynamic model under a single RGB-D camera setup. Still, the global texture map is static and cannot represent time-varying appearances of a dynamic object.

In this work, we propose a novel framework to generate a spatiotemporal texture map for a dynamic template model using a single RGB-D camera. Given a template model that is dynamically aligned with input RGB-D images, our main objective is to produce a spatiotemporal texture map that provides a time-varying texture of the model. Since a single camera can obtain only partial observations at each frame, we bring color information from other frames to complete a spatiotemporal texture map. 
We formulate a Markov random field (MRF) energy, where each node in the MRF indicates a triangular face of the dynamic template model at a frame. By minimizing the energy, for every frame, an optimal image frame to be used for texture mapping each face can be determined.
Recovering high-quality textures is another goal of this work because it is crucial for free-viewpoint video generation of a dynamic model. Due to imperfect motion tracking, texture drifts may happen 
when input color images are mapped onto the dynamic template model.
To resolve the problem, we find optimal texture coordinates for the template model on all input color frames. 
As a result, our framework can handle time-varying appearances of dynamic objects better than the previous approach~\cite{kim2019global} based on a single static texture map.

The key contributions are summarized as follows:
\begin{itemize}
    \item A novel framework to generate a stream of texture maps for a dynamic template model from color images captured by a single RGB-D camera.
    \item An effective MRF formulation including selective weighting to compose high-quality time-varying texture maps from single view observations.
    \item An accelerated approach to optimize texture coordinates for resolving texture drifts on a dynamic template model.
    \item Quantitative evaluation and user study to measure the quality of reconstructed time-varying texture maps.
\end{itemize}

\begin{figure*}[t]
	\centering
	\begin{tabular}{c@{\hspace{0.1cm}}c@{\hspace{0.1cm}}c}
		\includegraphics[width=1\textwidth]{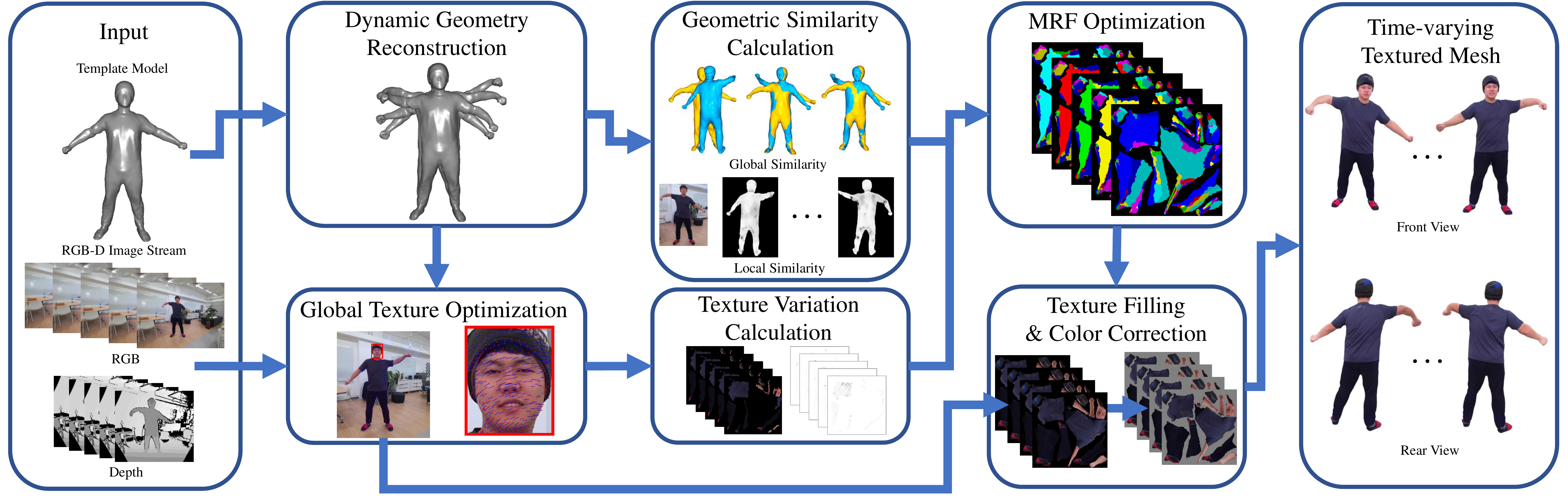}\\
	\end{tabular}
	\caption{System overview. Given a template model and a RGB-D image stream, we first reconstruct a dynamic object. We then optimize texture coordinates to resolve texture drifts among frames. Next, we calculate the geometric similarity and temporal texture variation, and formulate an MRF energy. By minimizing the MRF energy, frame labels are assigned to each face so that a high-quality spatiotemporal texture volume can constituted. As a post-processing, color correction resolves intensity differences among texture patches.}
	\label{fig:overallProcess}
\end{figure*}

\section{Related Work} 
\label{sec:Related_Work}

\subsection{Dynamic Object Reconstruction}

3D reconstruction systems for dynamic objects can be classified into two categories: template-based and templateless. A representative templateless work is DynamicFusion~\cite{newcombe2015dynamicfusion}, which shows real-time non-rigid reconstruction using a single depth camera. VolumeDeform~\cite{innmann2016volumedeform} considers a correspondence term made from SIFT feature matching to improve motion tracking. BodyFusion~\cite{BodyFusion}, which specializes in human body, exploits human skeleton information for surface tracking and fusion.

Template-based non-rigid reconstruction assumes that a template model for the target object is given, and deforms the template model to be aligned with input data. 
Most template-based methods~\cite{Li2009RobustSingle, Zollhofer2014, guo2015robust,Guo2018L0_tvcg} are based on embedded deformation (ED) graph model~\cite{Sumner2007Embedded}, which is also utilized in templateless works~\cite{3Dportraits, Dou_2015_CVPR, fusion4d, dou2017motion2fusion}.  
High-end reconstruction systems using a number of cameras can capture a complete mesh of the target object at each timestamp, so they commonly use template tracking between keyframes for compressing redundant data~\cite{fvv, Prada2017SpatioTemp, guo2019relightables}.

In this paper, we use a template-based method~\cite{ guo2015robust,Guo2018L0_tvcg} to track the motion of a dynamic object, while any method can be used if it can generate a sequence of meshes with a fixed connectivity conforming a RGB-D image sequence.

\subsection{Appearances of Reconstructed Models}

\paragraph*{Static scenes}
When scanning a static scene with a single RGB-D camera, 
color information obtained by volumetric fusion tends to be blurry due to inaccurate camera pose estimation.
Gal et al.~\cite{Seamless_Montage} optimize projective mapping from each face of a mesh onto one of the color images by performing combinatorial optimization on image labels and adjusting 2D coordinates of projected triangles.
Zhou and Koltun~\cite{ZhouColoMap} jointly optimize camera poses, non-rigid correction functions, and vertex colors to increase the photometric consistency.
Jeon et al.~\cite{jeon2016texture} optimize texture coordinates projected on input color images to resolve misalignments of textures.
Bi et al.~\cite{Bi2017bidirecti} propose patch-based optimization to align images against large misalignments between frames.
Fu et al.~\cite{fu2018texture} and Li et al.\cite{Li2019FastTexMapping_tvcg} improve the color of a reconstructed mesh by finding an optimal image label for each face. 
Fu et al.~\cite{fu2020joint} jointly optimize texture and geometry for better correction of textures. Huang et al.~\cite{huang2020adversarial} exploit a deep learning approach to reconstruct realistic textures.

\paragraph*{Non-rigid objects with multiple cameras}
Non-rigid object reconstruction has received attention in the context of performance capture using multi-view camera setup~\cite{fvv, Prada2017SpatioTemp, fusion4d, orts2016holoportation}. Fusion4D \cite{fusion4d} is a real-time performance capture system using multiple cameras, and it shows the capability to reconstruct challenging non-rigid scenes. However, it uses volumetrically fused vertex colors to represent the appearance, resulting in blurred colors of the reconstructed model. To resolve the problem, Holoportation~\cite{orts2016holoportation} proposes several heuristics, including color majority voting. Montage4D~\cite{du2018montage4d} considers misregistration and occlusion seams to obtain smooth texture weight fields.
Offline multi-view performance capture systems\cite{fvv, guo2019relightables} generate a high-quality texture atlas at each timestamp using mesh parameterization, and temporally coherent atlas parameterization \cite{Prada2017SpatioTemp} was proposed to increase the compression ratio of a time-varying texture map. 

\paragraph*{Non-rigid scenes with a single camera} 
Kim et al.~\cite{kim2019global} produces a global texture map for a dynamic object by optimizing texture coordinates on the input color images in a single RGB-D camera setup. Pandey et al.~\cite{pandey2019volumetric} propose semi-parametric learning to synthesize images from arbitrary novel viewpoints. PIFu~\cite{saito2019pifu} combines learning-based 3D reconstruction, volumetric deformation, and light-weight bundle adjustment to reconstruct a clothed human model in a few seconds from a single RGB-D stream.
These methods produce either a single texture map for the reconstructed dynamic model~\cite{kim2019global, saito2019pifu} or temporally independent blurry color images for novel viewpoints~\cite{pandey2019volumetric}. 
They all cannot generate a sequence of textures varying with motions of the reconstructed model over time. Our framework aims to produce a visually plausible spatiotemporal texture for a dynamic object with a single RGB-D camera.
\section{Problem Definition}
\label{sec:preliminary}

\subsection{Spatiotemporal Texture Volume}
Given a triangular mesh $\mathcal{M}$ of the template model, we assume that a non-rigid deformation function $\Psi_t$ is given for each time frame $t$, satisfying $\Psi_t(\mathcal{M})=\mathcal{M}_t$, where $\mathcal{M}_t$ denotes the deformed template mesh fitted to the depth map $D_t$. 
Our goal is to produce a spatiotemporal (time-varying) texture map $\mathcal{T}$ for the dynamic template mesh. By embedding $\mathcal{M}$ onto a 2D plane using a parameterization method \cite{UVatlas}, each triangle face $f$ of $\mathcal{M}$ is mapped onto a triangle in the 2D texture domain. A spatiotemporal texture map can be represented as stacked texture maps (or spatiotemporal texture volume), whose size is $W \times W \times T$, where $W$ is the side length of a rectangular texture map and $T$ is the number of input frames.
We assume that our mesh $\mathcal{M}_t$ keeps a fixed topology in the deformation sequence, and the position of face $f$ in the 2D texture domain remains the same regardless of time $t$. Only the texture information of face $f$ changes with time $t$ to depict dynamically varying object appearances.

\subsection{Objectives}
To determine the texture information for face $f$ of the deformed mesh $\mathcal{M}_{t}$ at time $t$ in a texture volume, we need to solve two problems: {\em color image selection} and {\em texture coordinate determination}.
We should first determine which color image in the input will be used for extracting the texture information. A viable option is to use the color image $\mathcal{C}_t$ at time $t$. However, depending on the camera motion, face $f$ may not be visible in $\mathcal{C}_t$, and in that case, texture should be borrowed from another color image $\mathcal{C}_t'$ where the face is visible. Moreover, when there are multiple such images, the quality of input images should be considered, as some images can provide sharper textures than others.

Once we have selected the color image $\mathcal{C}_{t'}$, we need to determine the mapped position of $f$ in $\mathcal{C}_{t'}$.
Since we already have the deformed mesh $\mathcal{M}_{t'}$ at time $t'$, the mapping of $f$ onto $\mathcal{C}_{t'}$ can be obtained by projecting face $f$ in $\mathcal{M}_{t'}$ onto $\mathcal{C}_{t'}$. However, due to the errors in camera tracking and non-rigid registration, mappings of $f$ onto different input color images may not be completely aligned. If we use such mappings for texture sampling, the temporal texture volume would contain little jitters in the part corresponding to $f$.

An overall pipeline of our system is shown in \Fig{overallProcess}. Starting from RGB-D images and a deforming model, we determine the aligned texture coordinates using global texture optimization~(\Sec{4_Initialization}). Then, to select color images for each face $f$ at all timestamps, we utilize MRF optimization (\Sec{5_method}). To reduce redundant calculation, we pre-compute and tabulate the measurements that are required to construct the cost function before MRF optimization. Finally, we build a spatiotemporal texture volume and refine textures by conducting post-processing.
\section{Spatiotemporal Texture Coordinate Optimization} 
\label{sec:4_Initialization}

Camera tracking and calibration errors induce misalignments among textures taken from different frames. Non-rigid registration errors incur additional misalignments in the case of dynamic objects.
Kim et al.~\cite{kim2019global} proposed an efficient framework for resolving texture misalignments for a dynamic object. 
For a template mesh $\mathcal{M}$, the framework optimizes the texture coordinates $\xi$ of mesh vertices on the input color images $\mathcal{C}$ so that sub-textures of each face extracted from different frames match each other. The optimization energy is defined to calculate the photometric inconsistency among textures taken from different frames: 
\begin{equation}
E(\xi,P)=\sum{||\mathcal{C}(\xi)-P||^2}, 
\label{equ:tex_cons}
\end{equation}
where $P$ denotes proxy colors computed by averaging $\mathcal{C}(\xi)$.
$\xi$ and $P$ are solved using alternating optimization, where $\xi$ is initialized with the projected positions $\Phi$ of the vertices of deformed meshes on corresponding color images.

To obtain texture coordinates with aligned textures among frames, we adopt the framework of Kim et al.~\cite{kim2019global} with some modification. Their approach selects keyframes to avoid using blurry color images (due to abrupt motions) in the image sequence and considers only those keyframes in the optimization. In contrast, we need to build a spatiotemporal texture volume that contains every timestamp, so optimization should involve all frames.
We modify the framework to reduce computational overhead while preserving the alignment quality. 

\begin{figure}[t]
	\centering
	\begin{tabular}{c@{\hspace{0.1cm}}c@{\hspace{0.1cm}}c}
		\includegraphics[width=0.15\textwidth]{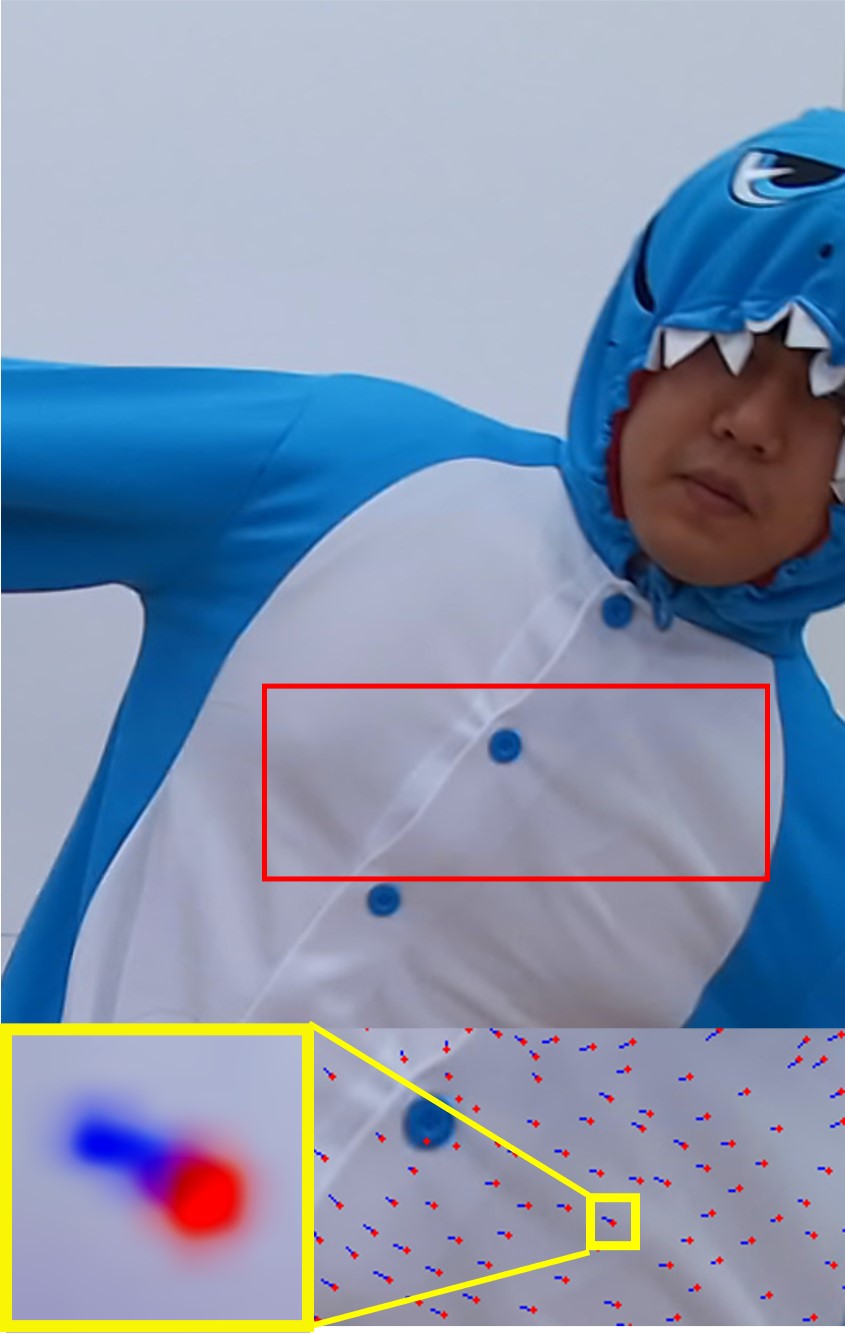} &
		\includegraphics[width=0.15\textwidth]{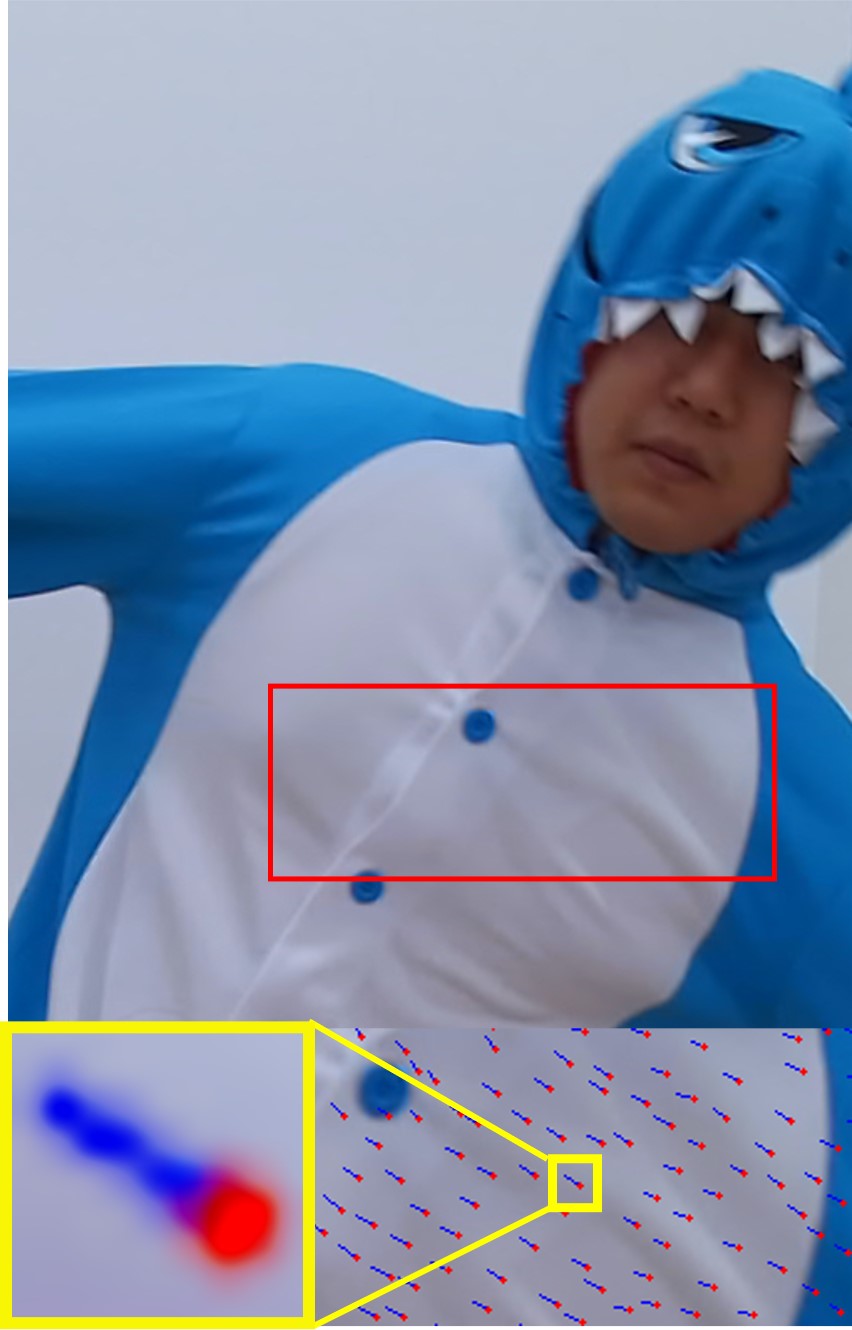} &
		\includegraphics[width=0.15\textwidth]{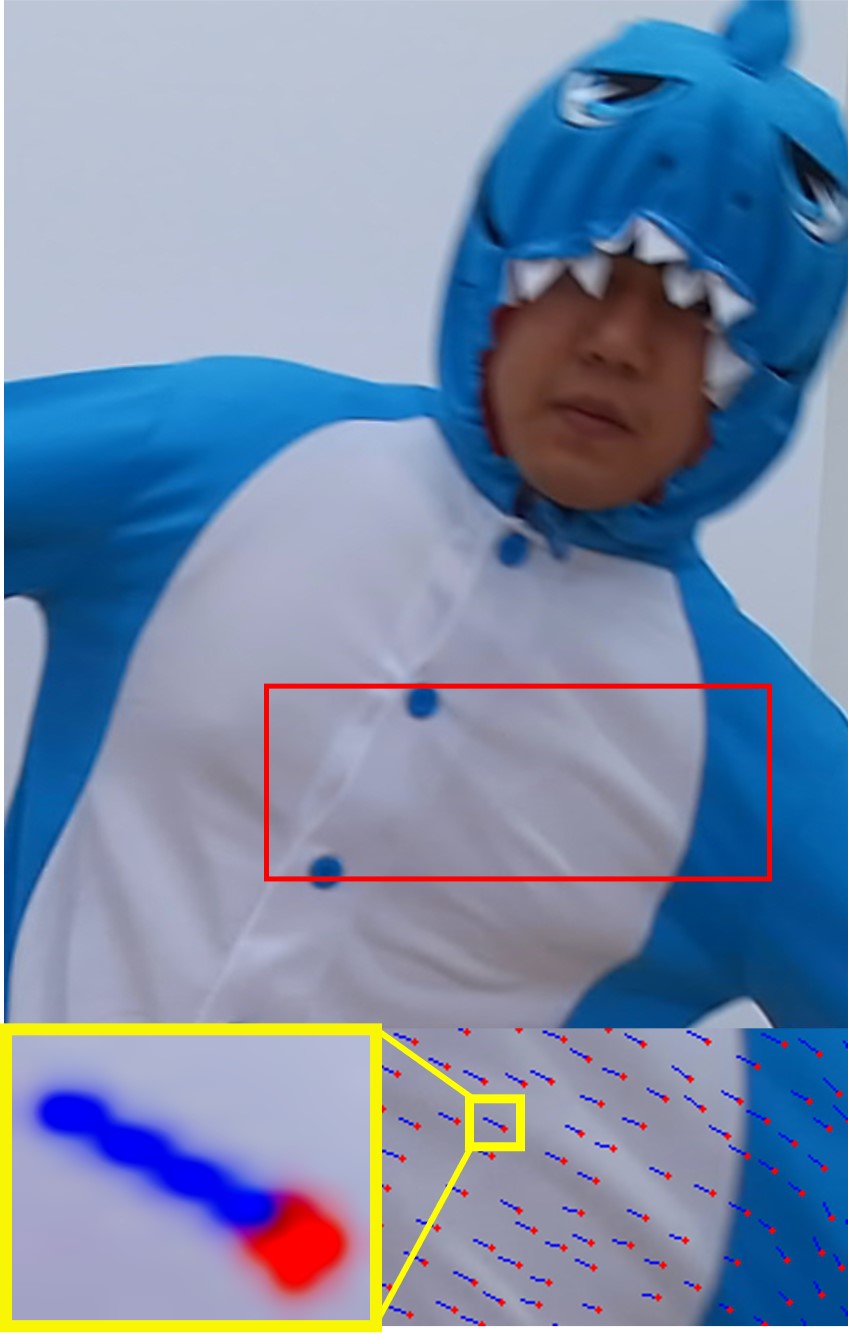}\\
		(a) $t-2$ & (b) $t$ & (c) $t+2$\\
	\end{tabular}
	\caption{Visualization of texture coordinate optimization. Due to tangential drifts, the displacements (blue lines) from the initial projected positions (red dots) to the optimized ones smoothly change over time.}
	\vspace*{-10pt}
	\label{fig:analysis[displacement]}
\end{figure}

Our key observation is that the displacements of texture coordinates determined by the optimization process smoothly change among adjacent frames. 
In~\Fig{analysis[displacement]}, red dots indicate initially projected texture coordinates $\Phi(V)$ of vertices $V$, and the opposite endpoints of blue lines indicate optimized texture coordinates $\xi(V)$. Based on the observation, we regularly sample the original input frames and use the sampled frames for texture coordinate optimization. Texture coordinates of the non-sampled frames are determined by linear interpolation of the displacements at the sampled frames.

In the optimization process, we need to compute the proxy colors $P$ from the sampled color images.
To avoid invalid textures (e.g., background colors), we segment foreground and background on the images using depth information with GrabCut~\cite{rother2004grabcut}, and give small weights for background colors.
After optimization with sampled images and linear interpolation for non-sampled images, most of the texture misalignments are resolved. Then, we perform a few iterations of texture coordinate optimization involving all frames for even tighter alignments.

Our modified framework is about five times faster than the original one~\cite{kim2019global}. In our experiment, where we sampled every four images from 
265 frames and the template mesh consists of 20k triangles, the original and our frameworks took 50 and 11 minutes, respectively.

\section{Spatiotemporal Texture Volume Generation} 
\label{sec:5_method}

After texture coordinate determination, we have aligned texture coordinates on every input image for each face of the dynamic template mesh.
The remaining problem is to select the color image $\mathcal{C}_{t'}$ for every face $f$ at each time frame $t$.
To this end, we formulate a labeling problem.

Let $\mathcal{L}_f$ be the candidate set of image frames for a face $f$, where $\mathcal{L}_f$ consists of image frames in which face $f$ is visible. Our goal is to determine the label $l_{f, t}$ from $\mathcal{L}_f$ for each face $f$ at each time $t$. If we have a quality measure for input color images, a direct way would be to select the best quality image in $\mathcal{L}_f$ for all $l_{f, t}$. However, this approach produces a static texture map that does not reflect dynamic appearance changes. 

To solve the labeling problem, we use MRF (Markov Random Field)~\cite{MRF_2, MRF_1} optimization.
The label $l_{f, t}$ for face $f$ at time $t$ is a node in the MRF graph, and there are two kinds of edges in the graph. A spatial edge connects two nodes $l_{f, t}$ and $l_{f', t}$ if faces $f$ and $f'$ are adjacent to each other in the template mesh. A temporal edge connects two nodes $l_{f, t}$ and $l_{f, t+1}$ in the temporal domain.
To determine the value of $l_{f, t}$ on the MRF graph, we minimize the following cost function.

\subsection{Cost Function}

Our cost function consists of data term $E_{d}$ and smoothness term $E_{s}$ as follows,
\begin{equation}
\begin{split}
    E(\mathcal{L}) = \sum_{f\in F}\sum_{t\in T}\Big({E_{d}(l_{f,t}) +\lambda\sum_{(f',t') \in \mathcal{N}_{f,t}} E_{s}(l_{f,t},l_{f',t},l_{f,t'})} \Big),
\end{split}
\end{equation}
where $F$ is the set of faces in the template mesh and $\mathcal{N}_{f,t}$ is the set of adjacent nodes of $l_{f,t}$ in the MRF graph.

\subsubsection{Data Term}
In many cases, a natural choice for the label of $l_{f, t}$ would be frame $t$. However, if face $f$ is not visible in $\mathcal{C}_{t}$, i.e., $t \notin \mathcal{L}_f$, we cannot choose $t$ for $l_{f, t}$. In addition, even in the case that $t \in \mathcal{L}_f$, $\mathcal{C}_{t}$ could be blurry and some other $\mathcal{C}_{t'}$, $t' \in \mathcal{L}_f$, could be a better choice.
To find a plausible label for $l_{f, t}$ in either case, we assume that similar shapes would have similar textures too. The data term consists of two terms $E_{qual}$ and $E_{geo}$:
\begin{equation}
    E_{data}(l_{f,t}) = E_{qual}(l_{f,t}) + E_{geo}(l_{f,t}) .
\end{equation}
$E_{qual}$ is designed to avoid assigning low-quality texture, such as blurred region. $E_{qual}$ is defined as follows:
\begin{equation}
\begin{split}
    E_{qual}(l_{f,t}) =&\mbox{ }
    \chi_{\theta_{b}}(|\xi(f, l_{f,t}) - \xi(f, l_{f,t}+1)|) \\
    & + \chi_{\theta_{n}}(1 - n_{f, l_{f,t}} \cdot c_{l_{f,t}}),
\end{split}
\label{eq:quality term}
\end{equation}
where $\chi_\theta(a)$ is a step function whose value is 1 if $a \geq \theta$, and 0 otherwise.
The first term estimates the blurriness of face $f$ by measuring the changes of the optimized texture coordinates $\xi$ of vertices among consecutive frames, $l_{f,t}$ and $l_{f,t}+1$.
The second term uses the dot product of the face normal $n$ and the camera's viewing direction $c$. $E_{qual}$ prefers to assign an image label if it provides sharp texture and avoids texture distortion due to slanted camera angle. By taking a step function $\chi_\theta$ in $E_{qual}$,
no penalty is imposed once the image quality is above the threshold, keeping the plausible label candidates as much as possible.

$E_{geo}$ ensures to select a timestamp with a similar shape. The geometry term measures global similarity $E_{glo}$ and local similarity $E_{loc}$ between shapes:
\begin{equation}
    E_{geo}(l_{f,t}) = \omega_{g} E_{glo}(l_{f,t}) + (1 - \omega_{g}) E_{loc}(l_{f,t}) .
\end{equation}
$E_{glo}$ considers overall geometric similarity between frames and is defined as follows:
\begin{equation}
    E_{glo}(l_{f,t}) = \frac{ \sum_p \mathcal{R}_{t \rightarrow l_{f,t}}(p)}{\Phi(K(\mathcal{M}_{t}), \mathcal{C}_{l_{f,t}}) \cup \Phi(\mathcal{M}_{l_{f,t}}, \mathcal{C}_{l_{f,t}})} ,
\label{equ:geo_global}
\end{equation}
where $p$ denotes an image pixel, and $\mathcal{R}_{t \rightarrow t'}(p) = \min \left(1, \left| \mathcal{\overline D}_{K(\mathcal{M}_{t})}(p) - \mathcal{\overline D}_{\mathcal{M}_{t'}}(p) \right|\right)$ is the clipped difference between two depth values $\mathcal{\overline D}_{K(\mathcal{M}_{t})}$ and $\mathcal{\overline D}_{\mathcal{M}_{t'}}$ that are rendered from $K(\mathcal{M}_{t})$ and $\mathcal{M}_{t'}$, respectively. Here, 
$K \in \mathbb{R}^{4\times4}$ denotes a 6-DoF rigid transformation that aligns $\mathcal{M}_{t}$ to $\mathcal{M}_{t'}$.

$E_{glo}$ is devised to find an image label $l_{f,t}$ such that the deformed template meshes $\mathcal{M}_t$ and $\mathcal{M}_{l_{f,t}}$ are 
geometrically similar.
It measures the sum of depth differences $\mathcal{R}$ over the union area of the projected meshes on image $\mathcal{C}_{l_{f,t}}$, $\Phi(K(\mathcal{M}_{t}), \mathcal{C}_{l_{f,t}}) \cup \Phi(\mathcal{M}_{l_{f,t}}, \mathcal{C}_{l_{f,t}})$, where $p$ in \Eq{geo_global} is a pixel in the union area on $\mathcal{C}_{l_{f,t}}$. We obtain $K$ with RANSAC based rigid pose estimation that utilizes coordinate pairs of the same vertices in $\mathcal{M}_{t}$ and $\mathcal{M}_{l_{f,t}}$.

Even though we found a geometrically similar frame ${l_{f,t}}$ with $E_{glo}$, local shapes around face $f$ do not necessarily match among $\mathcal{M}_{t}$ and $\mathcal{M}_{l_{f,t}}$.
Local  similarity $E_{loc}$ guides shape search using local geometric information and is defined as $E_{loc} = \left({-\frac{1}{3}\sum\tau_{j} \cdot \tau_{j'}}\right)$, where $(j, j')$ denotes one of three vertex pairs from face $f$ in $\mathcal{M}_{t}$ and $\mathcal{M}_{t'}$.
We utilize SHOT~\cite{tombari2010unique} to obtain a descriptor $\tau$ of the local geometry calculated for each vertex. SHOT originally uses the Euclidean distance to find the neighborhood set. Since we have a template mesh, we exploit the geodesic distance to obtain the local surface feature. To reduce the effect of noise, we apply median filter to SHOT descriptor values of vertices.

\subsubsection{Smoothness Term}
The smoothness term is defined for each edge of the MRF graph. Its role is to reduce seams in the final texture map and to control texture changes over time. To this end, we define spatial smoothness $E_{spa}$ in texture map domain and temporal smoothness $E_{temp}$ over time axis of the spatiotemporal texture volume. That is,
\begin{equation}
    E_{s}(l_{f,t},l_{f',t},l_{f,t'}) = \omega_{s}E_{spa}(l_{f,t},l_{f',t}) + \omega_{t}E_{temp}(l_{f,t},l_{f,t'}).
    \label{eq:smoothness}
\end{equation}
$E_{spa}$ is defined as follows:
\begin{equation}
\begin{split}
    E_{spa}(l_{f,t},l_{f',t}) = \frac{1}{|edge|} \sum_{p, p' \in edge}\Big( |&\mathcal{C}_{l_{f,t}}(p)-\mathcal{C}_{l_{f',t}}(p')| \\
    &+ |G_{l_{f,t}}(p)-G_{l_{f',t}}(p')| \Big),
\end{split}
\end{equation}
where $edge$ is a shared edge between two faces $f$ and $f'$, and 
$(p,p')$ are the corresponding image pixels when the edge is mapped onto $\mathcal{C}_{l_{f,t}}$ and $\mathcal{C}_{l_{f',t}}$. If the nodes $l_{f,t}$ and $l_{f',t}$ have different image labels, color information should match along the shared edge to prevent a seam in the final texture map.
In addition, we use the differences of gradient images, $G_{l_{f,t}}$ and $G_{l_{f',t}}$, to match the color changes around the edge between $f$ and $f'$.

$E_{temp}$ is defined as follows:
\begin{equation}
\begin{split}
    E_{temp}(l_{f,t},l_{f,t'}) &= \frac{1}{|face|} \sum_{p, p' \in face} |\mathcal{C}_{l_{f,t}}(p)-\mathcal{C}_{l_{f,t'}}(p')|,
\end{split}
\end{equation}
where $face$ denotes the projection of face $f$ onto $\mathcal{C}_{l_{f,t}}$ and $\mathcal{C}_{l_{f,t'}}$, and $(p,p')$ denotes corresponding image pixels on the projected faces. 
$E_{temp}$ measures the intensity differences and promotes soft transition of the texture assigned to $f$ over time shift $t\rightarrow t'$.

\begin{figure}[t]
	\centering
	\begin{tabular}{c@{\hspace{0.1cm}}c@{\hspace{0.1cm}}c}
		\includegraphics[width=0.15\textwidth]{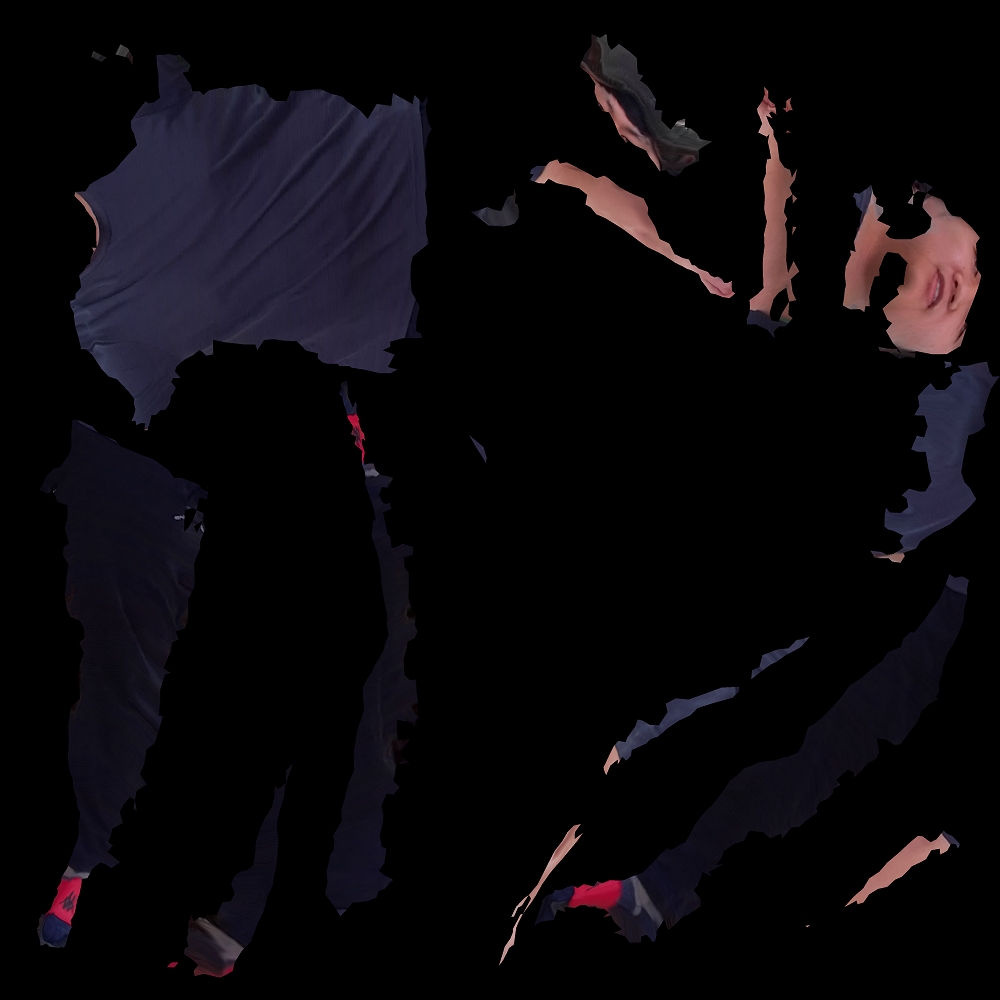} &
		\includegraphics[width=0.15\textwidth]{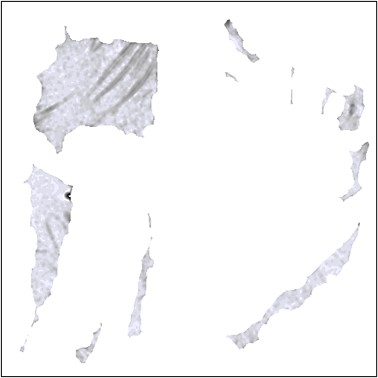} &
		\includegraphics[width=0.15\textwidth]{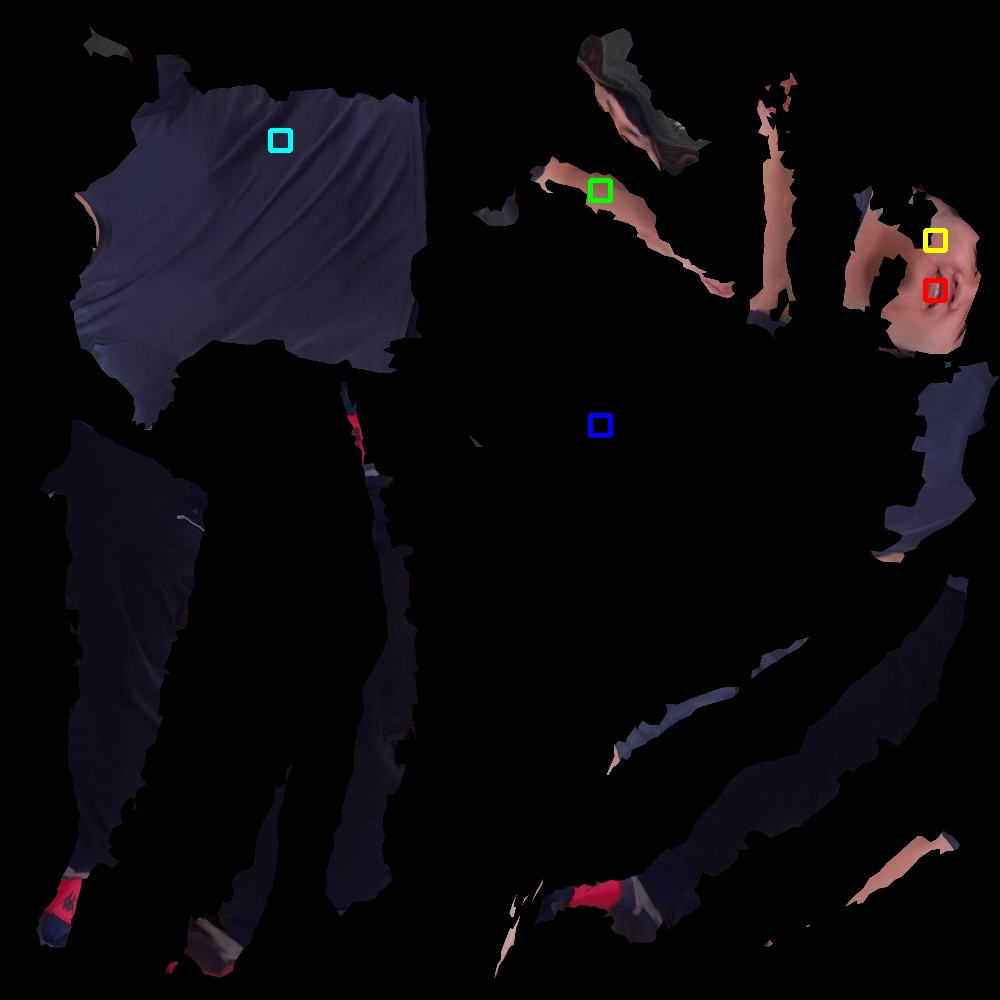}\\
		(a) & (b) & (c)\\
	\end{tabular}
	\begin{tabular}{c}
		\includegraphics[width=0.45\textwidth]{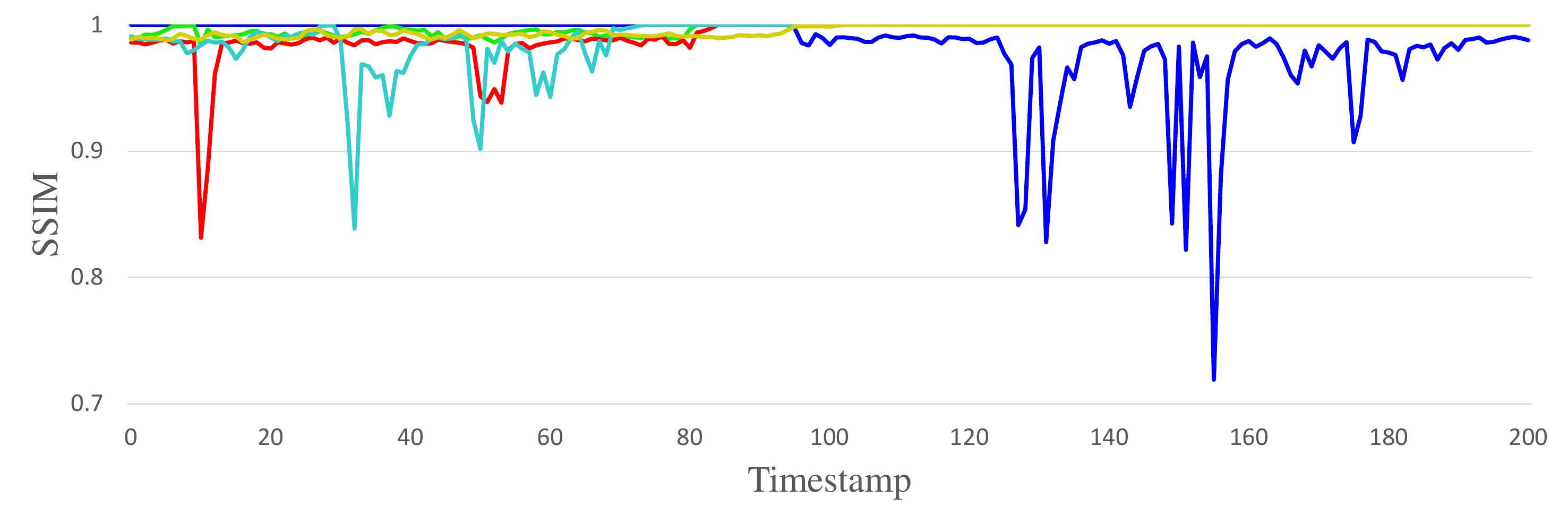}\\
		(d) Average SSIM values in colored boxes of (c) \\
	\end{tabular}
	\caption{Texture dynamism analysis. (a) texture map obtained by partial observation at timestamp $11$, (b) SSIM map between timestamps $11$ and $12$ (darker is smaller), (c) partial texture map at timestamp $12$. Blue and sky-blue boxes are on clothes whose appearances are frequently changing due to wrinkles, while green and yellow boxes are on consistent texture regions. Note that the texture under the blue box is not shown in (a) and (b), as it is invisible at timestamps $11$ and $12$. (d) The red curve corresponding to the red box on the mouth in (c) shows small SSIM values only at certain timestamps (e.g., opening and closing the mouth).}
	\vspace*{-25pt}
	\label{fig:components[SSIM]}
\end{figure}

\subsection{Selective Weighting for Dynamic Textures}

Textures on a dynamic object may have different dynamism depending on their positions. Some parts can have dynamic appearance changes with object motions, e.g., cloth wrinkles under human motions, while other parts may have almost no changes of textures over time, e.g., naked forearms of the human body.
We verified this property with an experiment.
After we have obtained optimized texture coordinates for a deforming mesh sequence, we can build an initial incomplete spatiotemporal texture volume by simply copying the visible parts in the input images onto the corresponding texture maps (\Fig{components[SSIM]}a).
Then, we compute the SSIM map between adjacent frames in the texture volume on the overlapping areas (\Fig{components[SSIM]}b).
\Fig{components[SSIM]}d shows changes in the SSIM values over time for the marked parts in \Fig{components[SSIM]}c. In the blue and cyan colored regions, SSIM fluctuates along time, meaning that these parts have versatile textures. Such parts need to be guaranteed for dynamism in the final spatiotemporal texture, while smooth texture changes would be fine for other parts.

To reflect this property in the MRF optimization, we modify the smoothness terms on MRF edges.
We ignore a temporal MRF edge if the labels of the linked nodes exhibit a lower SSIM value, i.e., less correlation.
In our temporal selective weighting scheme, the following new temporal smoothness term replaces $E_{temp}(l_{f,t},l_{f,t'})$ in Eq.~(\ref{eq:smoothness}):
\begin{equation}
\begin{split}
    E_{temp}&(l_{f,t},l_{f,t'}) \leftarrow \\ &E_{temp}(l_{f,t},l_{f,t'})\cdot\chi_{\theta_{\Omega}}\Big(\min\big(S(f,l_{f,t}),S(f,l_{f,t'})\big)\Big) ,
\end{split}
\label{equ:revised_temp}
\end{equation}
where $S(f, \cdot)$ is the average SSIM value of the pixels inside face $f$ computed among consecutive frames. $\chi_{\theta_\Omega}(a)$ is a step function whose value is 1 if $a \geq \theta_\Omega$, and 0 otherwise.
This scheme encourages the dynamism of textures over time when there is less correlation among frames.

In a similar way, we update the spatial smoothness term $E_{spa}(l_{f,t},l_{f',t})$ in Eq.~(\ref{eq:smoothness}) as follows:
\begin{equation}
\begin{split}
    E_{spa}&(l_{f,t},l_{f',t}) \leftarrow \\
    &E_{spa}(l_{f,t},l_{f',t}) \Bigg( 2 - \frac{\left(\chi_{\theta_{\Omega}}\big(S_{m}(f,k)\big) + \chi_{\theta_{\Omega}}\big(S_{m}(f',k)\big)\right)}{2} \Bigg),
\end{split}
\end{equation}
where $S_{m}(f,k)=\min\limits_{k \in \mathcal{L}_f}(S(f,k))$.
If there is a label with a small SSIM value in $\mathcal{L}_f$, the part has a risk of temporal discontinuity due to the updated temporal smoothness term in \Eq{revised_temp}.
Therefore, in that case, we enforce the spatial smoothness term to 
impose spatial coherence when temporal discontinuity happens.
In \Fig{components[SSIM]}c, the red box is likely to exhibit temporal changes, e.g., mouth opening. 
Our spatial selective weighting scheme increases the spatial smoothness term for the red box to secure spatially coherent changes of texture.

\subsection{Labeling and Post Processing}

Since our MRF graph is large to solve, we divide the problem and solve it using alternating optimization. The idea of fixing a subset and optimizing the others is known as block coordinate descent (BCD). In~\cite{chen2014fast, thuerck2016fast}, the BCD technique is applied to uniform and non-uniform MRF topology.
Our MRF graph consists of spatial edges connecting adjacent mesh faces at a single frame and temporal edges connecting corresponding mesh faces at consecutive frames.
As a result, our MRF topology is spatially non-uniform but temporally uniform. We subdivide the frames into two sets: even-numbered and odd-numbered. We first conduct optimization for even-numbered frames without temporal smoothness terms. After that, we optimize for odd-numbered frames while fixing the labels of even-numbered frames, where temporal smoothness terms become unary terms. We iterate alternating optimization among even- and odd-numbered frames until convergence. To optimize a single set MRF graph (even or odd), we use an open-source library released by Daniel et al.~\cite{thuerck2016fast}. 

\begin{figure}[t]
	\centering
	\begin{tabular}{c@{\hspace{0.1cm}}c@{\hspace{0.1cm}}c}
		\includegraphics[width=0.14\textwidth]{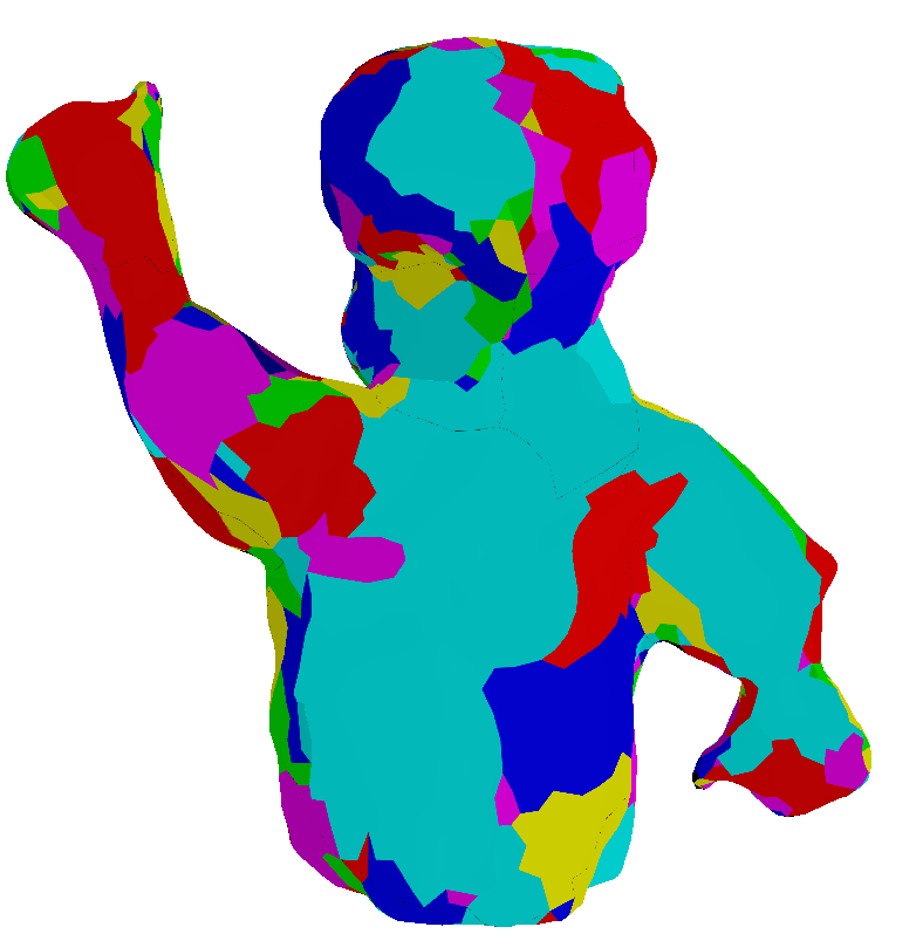} &
		\includegraphics[width=0.14\textwidth]{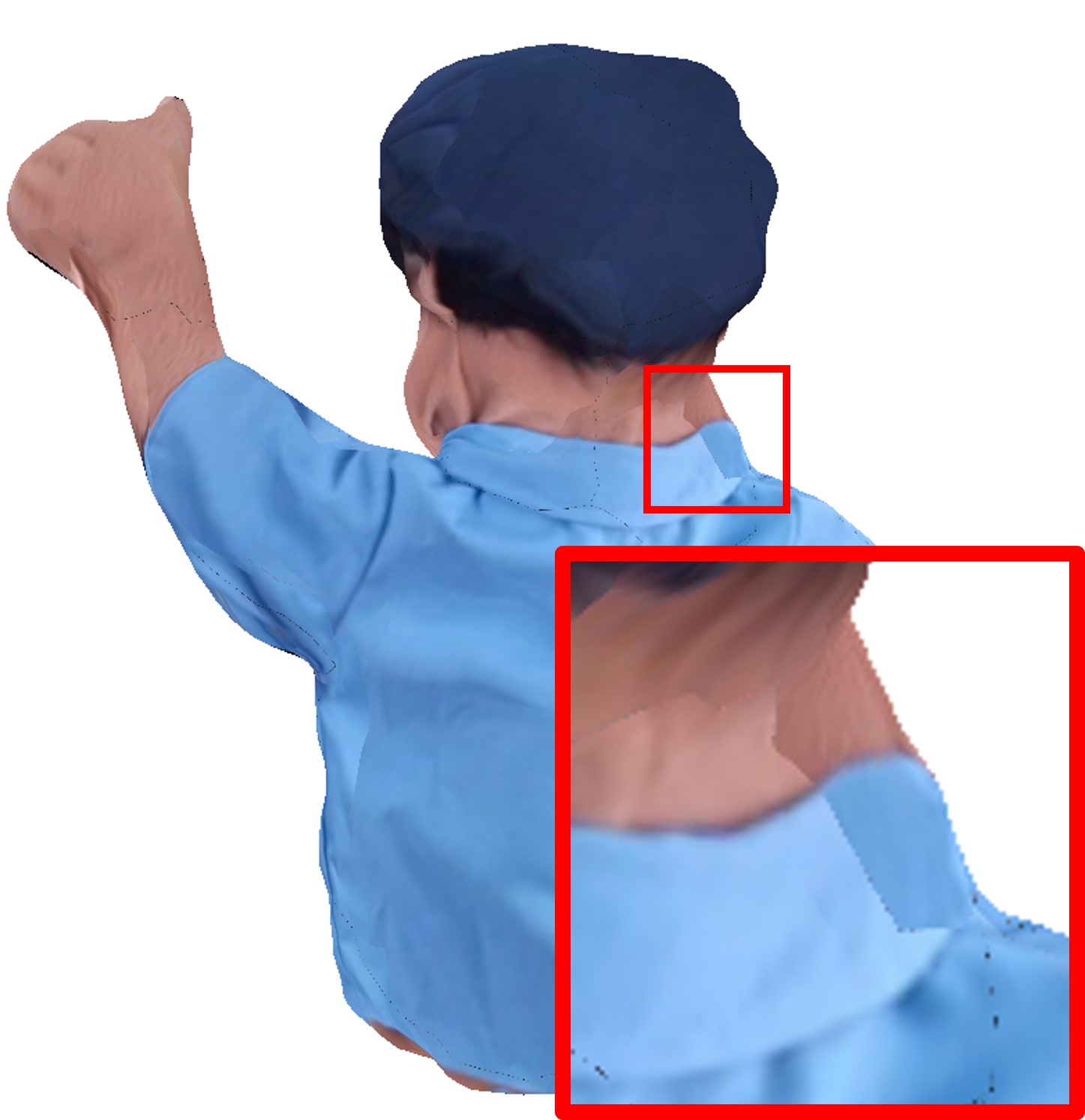} &
		\includegraphics[width=0.14\textwidth]{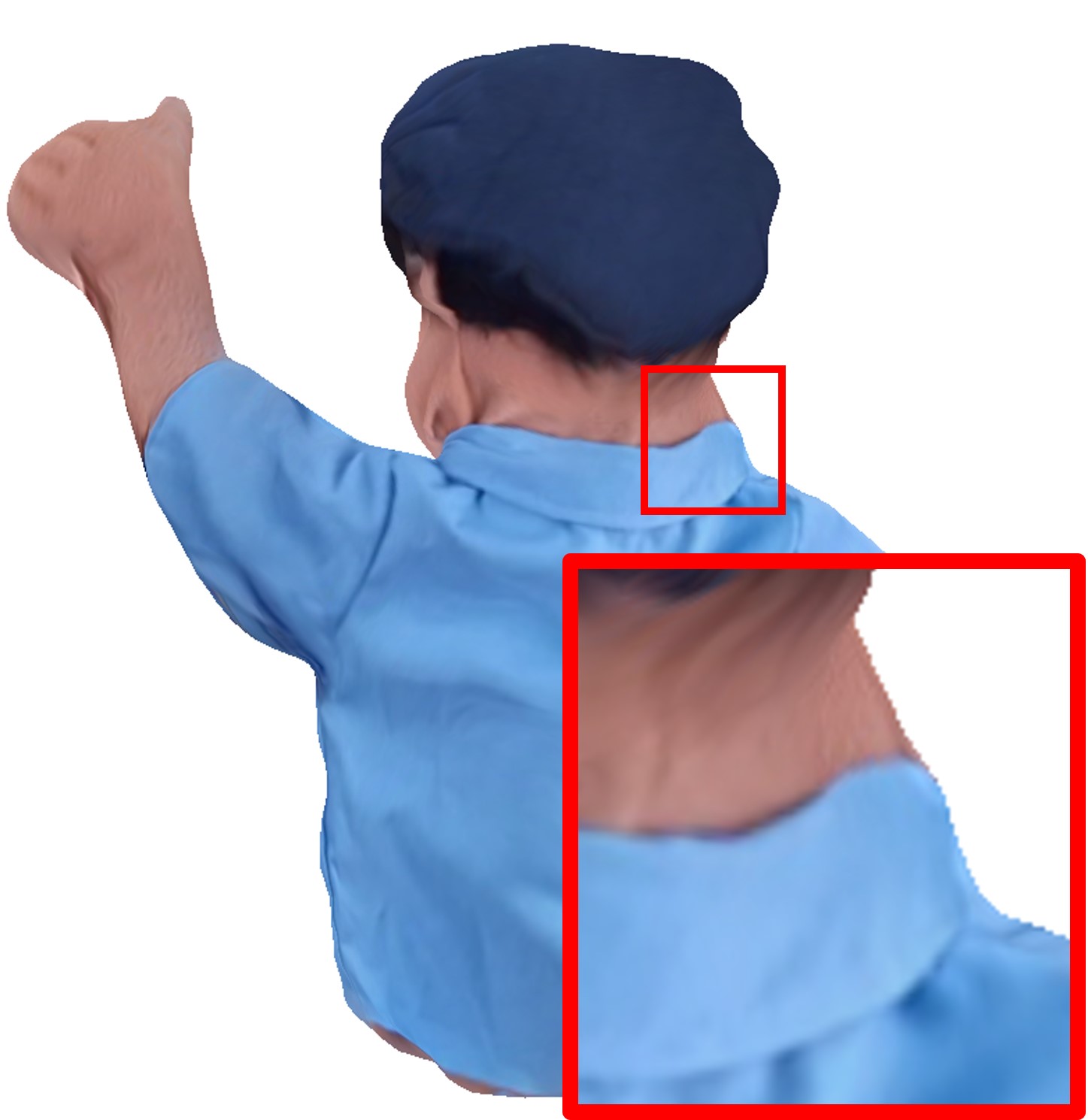}\\
		(a) & (b) & (c)\\
	\end{tabular}
	\caption{Effect of color correction. (a) color coded frame labels overlaid on the mesh. (b) before color correction. (c) after color correction.}
	\label{fig:analysis[colorcorrection]}
	\vspace*{-12pt}
\end{figure}

After labeling is done, we have a frame label per face at each timestamp. Then we fill the spatiotemporal texture atlas volume by copying texture patches from the labeled frames. 
When there are drastic brightness changes among different frames,
spatial seams may happen on label boundaries (\Fig{analysis[colorcorrection]}a). 
To handle such cases,
we apply gradient processing~\cite{prada2018gradient} to texture slices in the volume,
resolving spatial seams (\Fig{analysis[colorcorrection]}c).

\begin{figure*}[t]
	\centering
	\begin{tabular}[t]{l@{\hspace{0.1cm}}c@{\hspace{0.2cm}}c}
		\specialcell[b]{w/o temporal\\smoothness}&
        \includegraphics[width=0.35\textwidth]{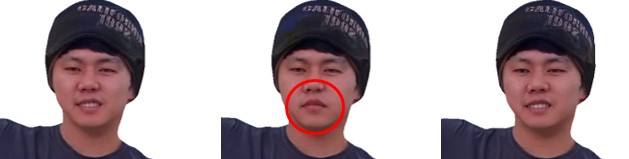} &
		\includegraphics[width=0.45\textwidth]{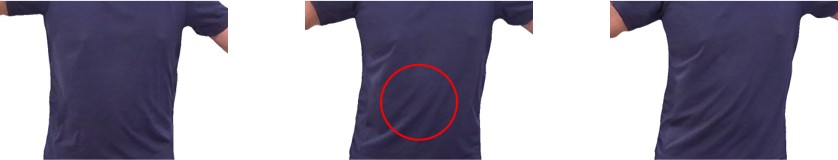}\\
		\specialcell[b]{w/ constant\\weighting}&
        \includegraphics[width=0.35\textwidth]{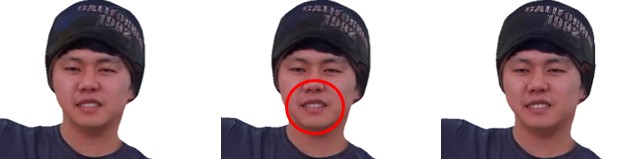} &
		\includegraphics[width=0.45\textwidth]{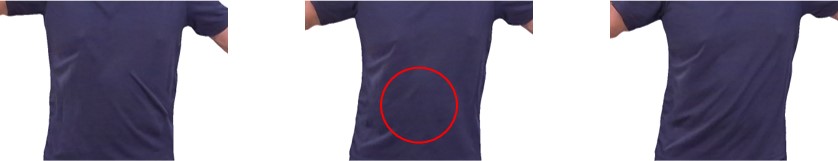}\\
		\specialcell[b]{w/ selective\\weighting}&
        \includegraphics[width=0.35\textwidth]{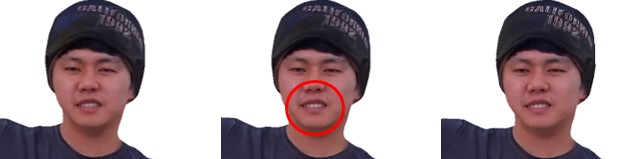} &
		\includegraphics[width=0.45\textwidth]{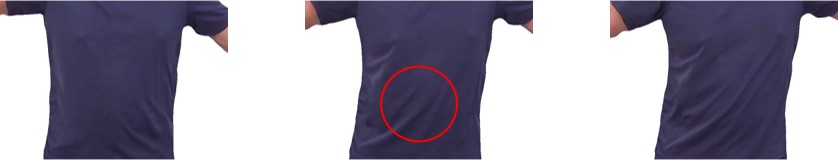}\\
		& (a) mouth changes & (b) wrinkle changes \\ 
	\end{tabular}
	\vspace{-1mm}
	\caption{Effect of selective temporal weighting. 
	 Each row of (a) and (b) shows texture changes in three consecutive frames.
	 (top) Without the temporal smoothness term, dynamism is guaranteed because there are no restrictions. However, some areas may exhibit unnatural changes (e.g., mouth opened and closed in consecutive frames). (middle) Uniform temporal smoothness term produces the opposite effect (no dynamism). (bottom) Selective weighting distinguishes parts that need dynamism and produces natural results.}
	\label{fig:ablation[temporal selective weighting]}
\end{figure*}

\begin{figure}[t]
	\centering
	\begin{tabular}{c@{\hspace{0.1cm}}c}
		\includegraphics[width=0.22\textwidth]{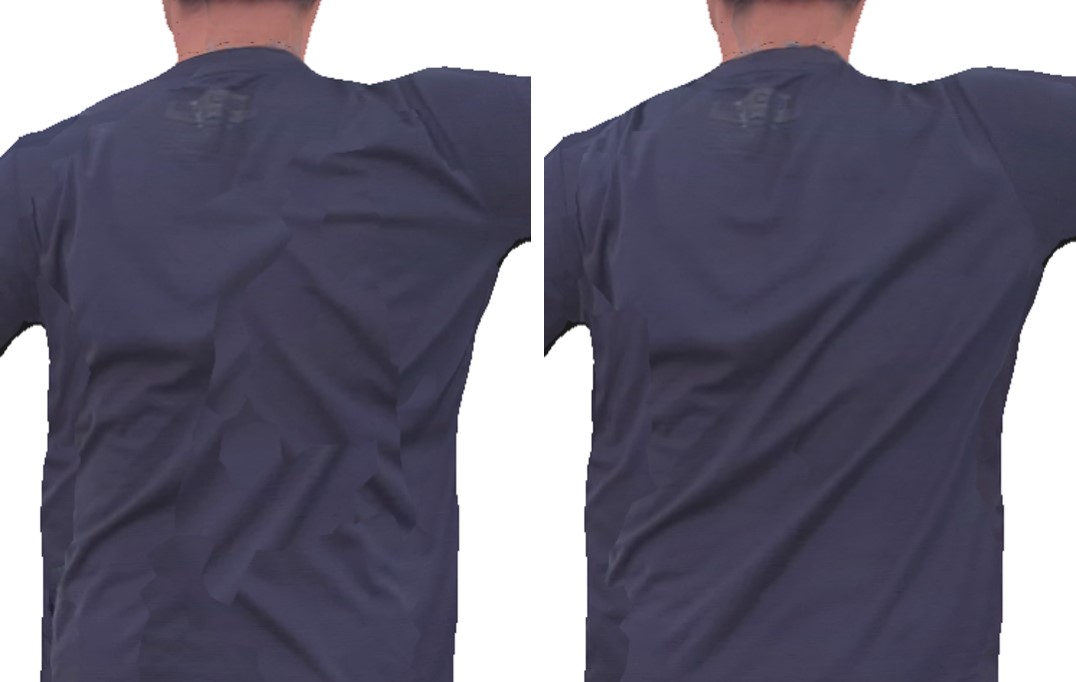} &
		\includegraphics[width=0.235\textwidth]{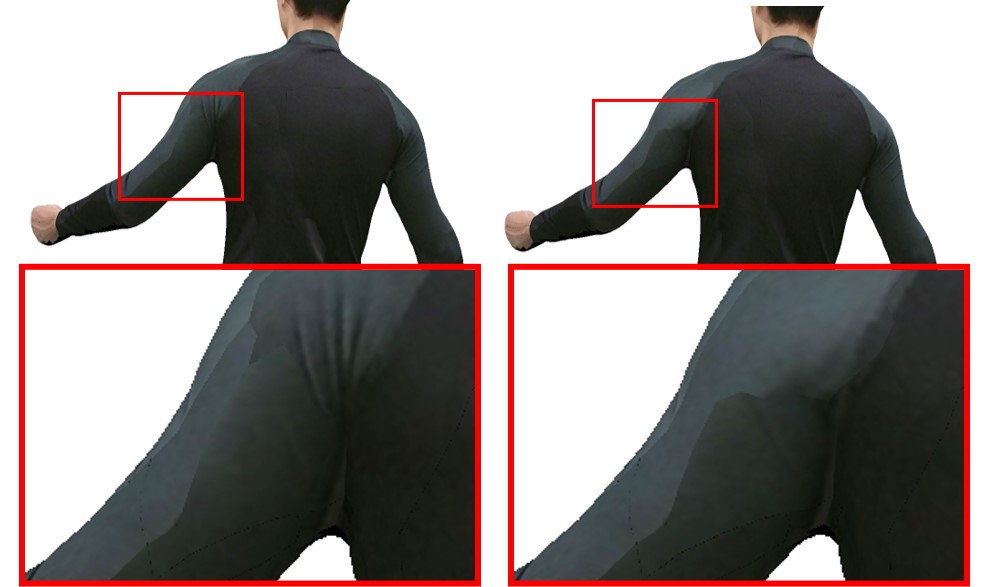} \\
		(a)  & (b)\\
	\end{tabular}
	\caption{Effects of geometric similarity terms. (a) Global similarity (right) helps reconstruct proper and consistent texture in a large region, compared to using only local similarity (left). 
	(b) Only using global similarity (left) may borrow texture from an overall similar frame ignoring local geometry. Combining local similarity additionally (right) enables finding more appropriate texture from one of similar frames.
	}
	\label{fig:ablation[global and local similarity]}
\end{figure}

\section{Results} 
\label{sec:6_results}

\subsection{Experiment Details}

\paragraph*{Setup}
All datasets shown in this paper were recorded by an Azure Kinect DK~\cite{KinectAzure} with 3072p resolution for color images, and $640 \times 576$ for depth images. All experiments were conducted on a workstation equipped with an Intel i7-7700K 4.2GHz CPU, 64GB RAM, and NVIDIA Titan RTX GPU. Our RGB-D sequences used for experiments contain 200 to 500 frames. 
The size of a texture slice in a spatiotemporal texture volume is $4000 \times 4000$.
We set $\omega_g=0.9$, $\omega_s=10$, $\omega_t=2$, $\theta_b=20$, $\theta_n=0.3$, $\theta_{\Omega}=0.95$ for all examples.
Source code for our method can be found at https://github.com/posgraph/coupe.Sptiotemporal-Texture/.

\begin{table}[ht]
    \centering
    \begin{tabular}{ c || c | c | c | c | c | c}
        \hline
        \hline
        Scene & 1 & 2 & 3 & 4 & 5 & 6 \\
        \hline
        \# Frame & 265 & 210 & 372 & 247 & 484 & 210 \\
        \hline
        Tex Opt (min) & 11.0 & 9.3 & 8.8 & 5.6 & 19.2 & 8.6 \\
        \hline
        Labeling (min) & 23.0 & 17.9 & 20.0 & 23.9 & 56.0 & 25.1 \\
        \hline 
        
    \end{tabular}
    \caption{Computation times. `Tex Opt' means global texture coordinate optimization. The input images of scenes 3 and 4 are of lower resolutions than those of other scenes.}
    \label{tbl:timing}
    \vspace*{-12pt}
\end{table}

\paragraph*{Timing}
The global texture coordinate optimization takes about $9 \sim 20$ minutes for a dynamic object. Solving MRF on average takes about $5 \sim 7$ seconds per frame. The color correction takes about 20 seconds per frame, but this step can be parallelized. Timing details are shown in \Tbl{timing}.

\paragraph*{Dynamic object reconstruction}
We reconstruct a template model by rotating a color camera around the motionless object
and using Agisoft Metashape\footnote{https://www.agisoft.com/} for 3D reconstruction. 
Then we apply mesh simplification~\cite{DOD2006} to the reconstructed template model and reduce the number of triangle faces to about 10$\sim$30k. 
To obtain a deformation sequence of the template mesh $\mathcal{M}$ that matches the input depth stream,
we adopt a non-rigid registration scheme based on an embedded deformation (ED) graph \cite{Sumner2007Embedded}.
The scheme parameterizes a non-rigid deformation field using a set of 3D affine transforms. It estimates the deformation field to fit mesh $\mathcal{M}$ into an input depth map $D_t$ at time $t$ using an $l_0$ motion regularizer \cite{guo2015robust}.

\subsection{Analysis}

\paragraph*{Texture coordinate optimization}
To accelerate the texture coordinate optimization process, we regularly sample frames and interpolate the texture coordinate displacements computed from the samples. \Tbl{tco_sampling} shows that our approach achieves almost the same accuracy compared to the original one using all frames. Using fewer sample frames reduces more computation, and we sample every four frames for the time-accuracy tradeoff.

\begin{table}[ht]
    \centering
    \begin{tabular}{ c || c | c | c | c }
        \hline
        sampling & 1/2 & 1/3 & 1/4 & 1/5 \\
        \hline
        error ratio & 1.0063 & 1.02220 & 1.02309 & 1.0427 \\
        \hline 
    \end{tabular}
    \caption{Error ratio of our modified texture coordinate optimization depending on the sampling ratio, where 1/$n$ means sampling every $n$ frames. We used the value of \Eq{tex_cons} to measure the error of a texture coordinate optimization result. The error ratio was computed by dividing the error of our sampling approach by that of the original approach, taking the average of all datasets.}
    \label{tbl:tco_sampling}
\end{table}

\paragraph*{Global \& local geometric similarity}
Our method utilizes global and local geometric similarities to assign the most reasonable frame labels to the faces of the dynamic template mesh. In \Fig{ablation[global and local similarity]}a, back parts that do not show obvious geometric patterns are not distinguishable by local similarity, and global similarity helps to assign proper frame labels. On the other hand, as shown in \Fig{ablation[global and local similarity]}b, local similarity helps to find appropriate textures if local geometry changes should be considered to borrow natural appearances from other frames.

\begin{figure}[t]
	\centering
	\begin{tabular}{c@{\hspace{0.1cm}}c}
		\includegraphics[width=0.15\textwidth]{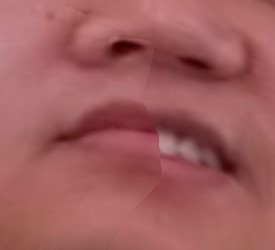} &
		\includegraphics[width=0.15\textwidth]{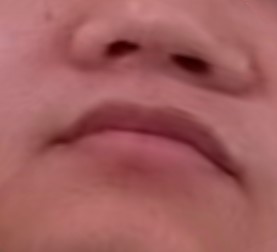} \\
		(a)  & (b)\\
	\end{tabular}
\caption{Without (a) and with (b) spatial selective weighting. With spatial selective weighing, coherent textures are obtained in regions where textures may change dynamically.}
	\label{fig:ablation[spatial selective weighting]}
\end{figure}

\begin{table*}[h]
    \centering
    \begin{tabular}{| c | c | c | c | c | c | c | c |}
        \hline
        Metrics & Methods & Scene 1 & Scene2 & Scene 3 & Scene 4 & Scene 5 & Scene 6 \\
        \hline
        \hline
        \multirow{2}{*}{\shortstack{PSNR \\ (every frame)}} & Kim et al.~\cite{kim2019global} & 28.1677 & 29.3667 & 26.5816 & 27.1487 & 21.8453 & 27.2445 \\  
        \cline{2-8} 
        & Ours & 28.6900 & 30.6023 & 28.9535 & 28.0716 & 23.4724 & 27.3731 \\
        \hline
        \hline 
        \multirow{4}{*}{\shortstack{PSNR \\ (unseen frame (1/6))}} 
        & Kim et al.~\cite{kim2019global} & 28.1243 & 29.2732 & 26.6415 & 27.1501 & 21.6907 & 26.8947\\  
        \cline{2-8} 
        & Ours (1/3) & 28.8495 & 29.9831 & 28.4498 & 27.6200 & 23.1869 & 27.2290\\  
        \cline{2-8} 
        & Ours (1/2) & 28.8797 & 30.0269 & 28.5851 & 27.8092 & 23.2763 & 27.2566\\  
        \hline
        \hline
        \multirow{3}{*}{\shortstack{Blurriness \\ (Crete et al.~\cite{crete2007blur})}} 
        & Input images & 0.7937 & 0.8004 & 0.8906 & 0.9095 & 0.8220 & 0.8440 \\
        \cline{2-8}
        & Kim et al.~\cite{kim2019global} & 0.7971 & 0.8006 & 0.8865 & 0.9101 & 0.8384 & 0.8514 \\   
        \cline{2-8}
        & Ours & 0.7896 & 0.7912 & 0.8824 & 0.9065 & 0.8104 & 0.8425 \\
        \hline 
    \end{tabular}
    \caption{Texture reconstruction errors and blurriness. PSNR and blurriness were measured excluding the background. For blurriness, smaller means sharper. In the top rows, average PSNR was computed using all frames. In the middle rows, 1/$n$ in parentheses means sampling every $n$ frames, and average PSNR was computing using 1/6 sample frames that were not involved in texture reconstruction.
    }
    \label{tbl:result_quan}
\end{table*}
\begin{figure*}[t]
	\centering
	\begin{tabular}{c@{\hspace{0.1cm}}c@{\hspace{0.1cm}}c}
		\includegraphics[width=0.58\textwidth]{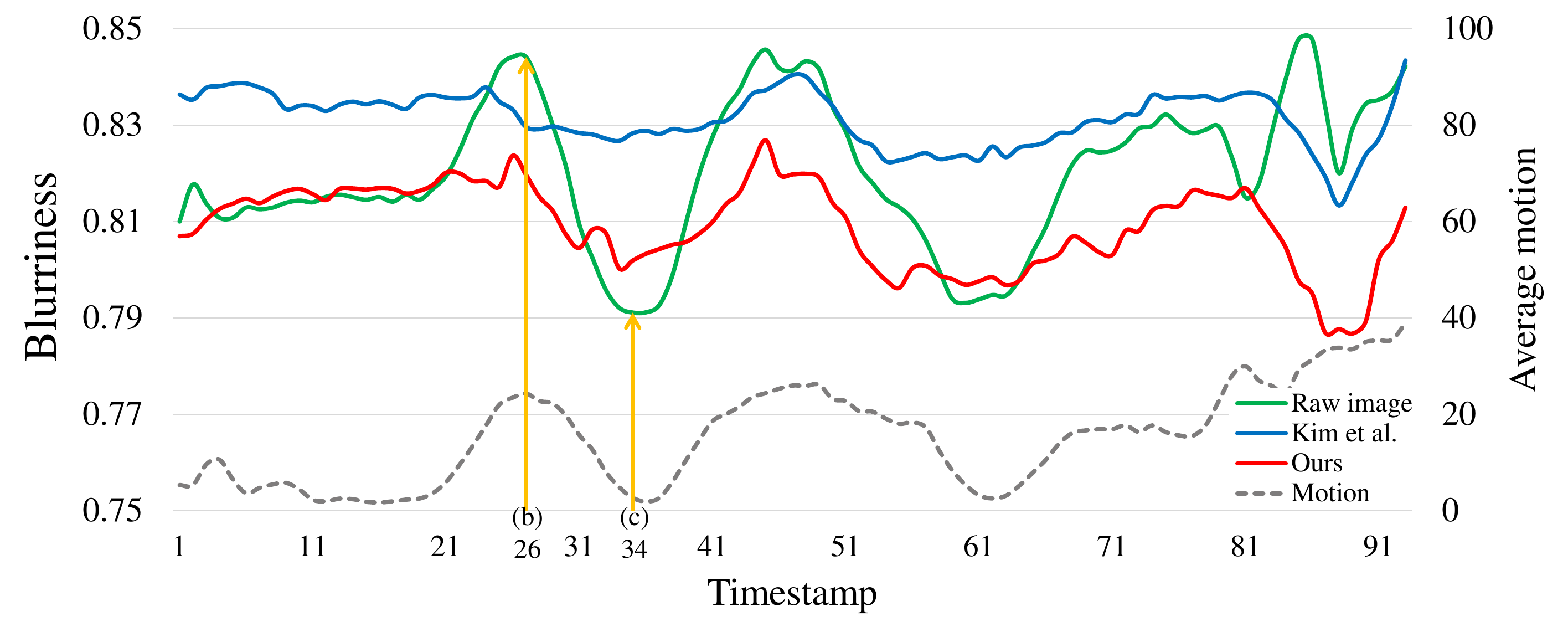} &
		\includegraphics[width=0.194\textwidth]{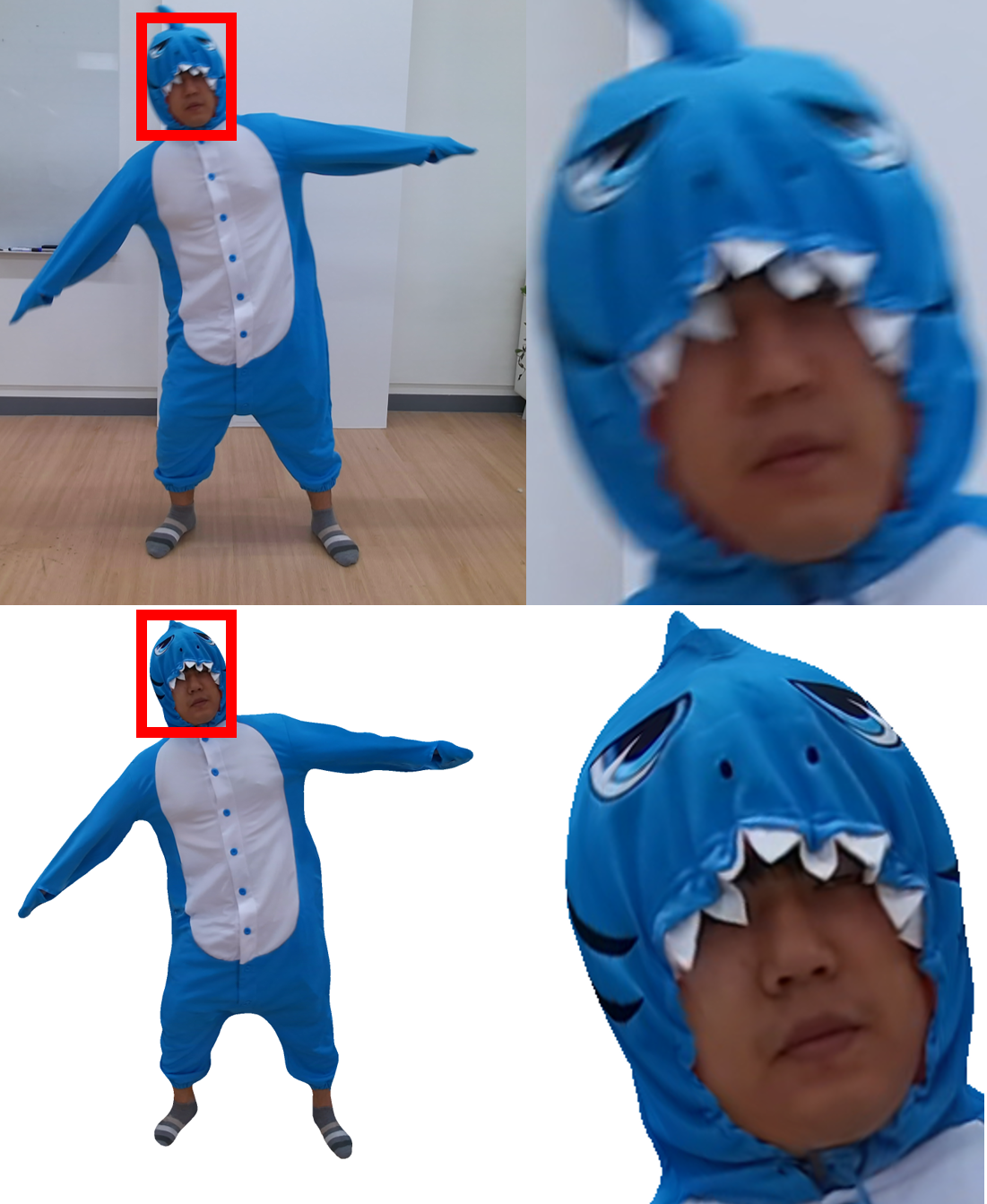} &
		\includegraphics[width=0.195\textwidth]{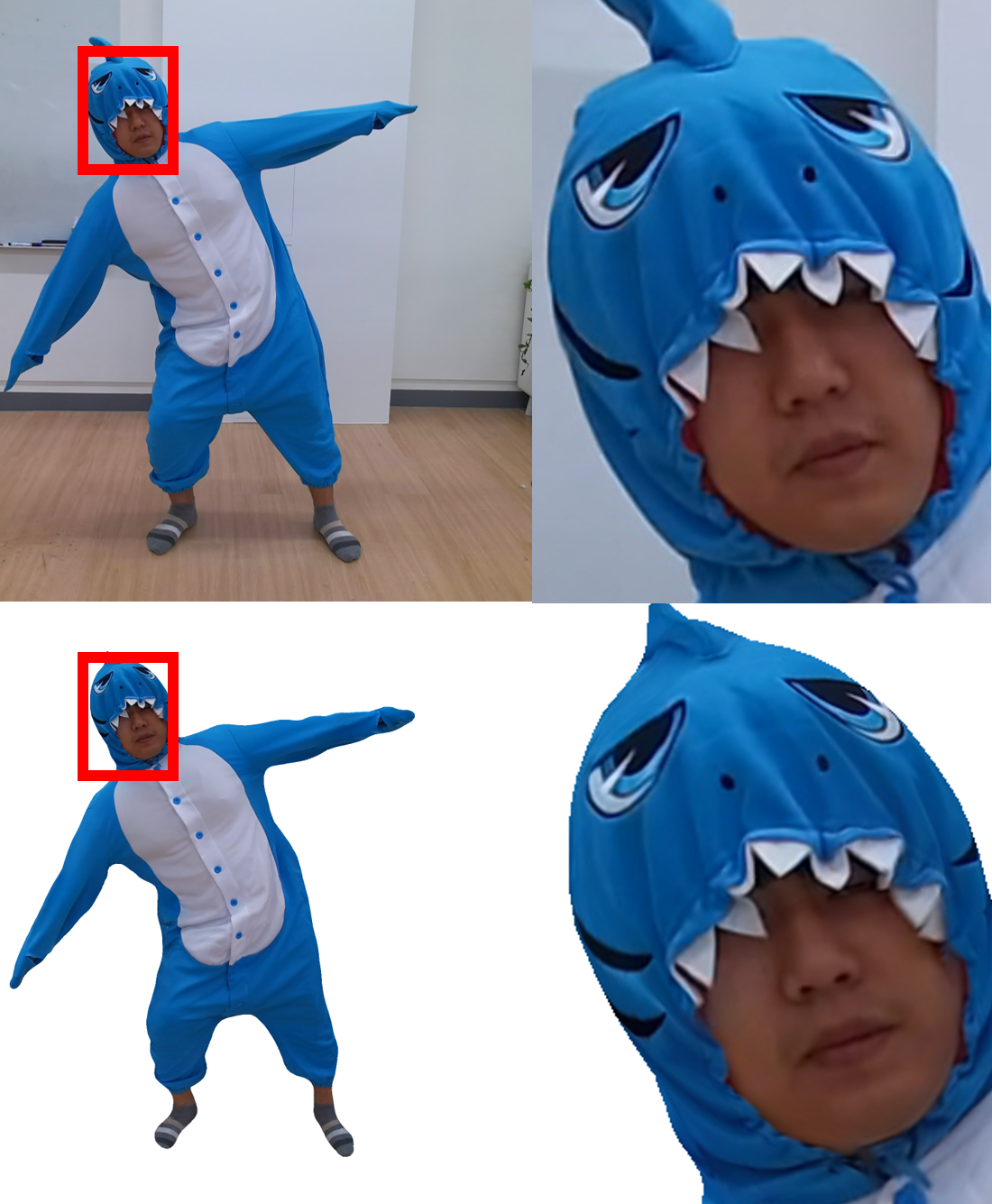}\\
		(a) & (b) frame 26 & (c) frame 34
	\end{tabular}
	\caption{Texture reconstruction quality. (a) the blurriness of input and reconstructed textures as well as average vertex motions of frames. (b)(c) input (top) and rendered (bottom) images. The reconstructed texture is apparent in (b), although the input is blurry due to fast motion. In (c), the reconstructed texture preserves the sharpness of the input.}
	\label{fig:quan[blur]}
\end{figure*}

\paragraph*{Selective weighting}
\Fig{ablation[spatial selective weighting]} shows that spatial selective weighting can help preventing abnormal seams. \Fig{ablation[temporal selective weighting]} compares different temporal weighting schemes. Our temporal selective weighting makes the mouth motion consistent (\Fig{ablation[temporal selective weighting]}a), and incurs more dynamism on the back part by dropping the smoothness constraint (\Fig{ablation[temporal selective weighting]}b).

\subsection{Quantitative Evaluation}
\label{sec:quantitative}

\paragraph*{Coverage}
We computed the amount of textures generated by our method. Naive projection can cover 29.5\% $\sim$ 39.6\% of the mesh faces depending on the scene, while our reconstructed textures always cover the entire surfaces.

\paragraph*{Reconstruction error}
We quantitatively measured the reconstruction errors by calculating PSNR between input images and rendered images of the reconstructed textures. To circumvent slight misalignments, we subdivided the images into small grid patches and translated the patches to find the lowest MSE. The top rows of \Tbl{result_quan} show that our spatiotemporal textures reconstruct the input textures more accurately than global textures~\cite{kim2019global}.
Even when intentionally we did not use some input images (e.g., every even frames) for texture reconstruction, our method could successfully produce dynamic textures, as shown in the middle rows of \Tbl{result_quan}. 

\paragraph*{Texture quality}
We quantitatively assessed the quality of reconstructed textures using a blur metric~\cite{crete2007blur}. The bottom rows of \Tbl{result_quan} show our results are sharper than global textures~\cite{kim2019global}. Our results are even sharper than the input images on average, as texture patches containing severe motion blurs are avoided in our MRF optimization. See \Fig{quan[blur]} for illustration.

\subsection{Qualitative Comparison}

\begin{figure*}[!t]
	\centering
	\begin{tabular}{c@{\hspace{0.1cm}}c}
		\includegraphics[width=0.48\textwidth]{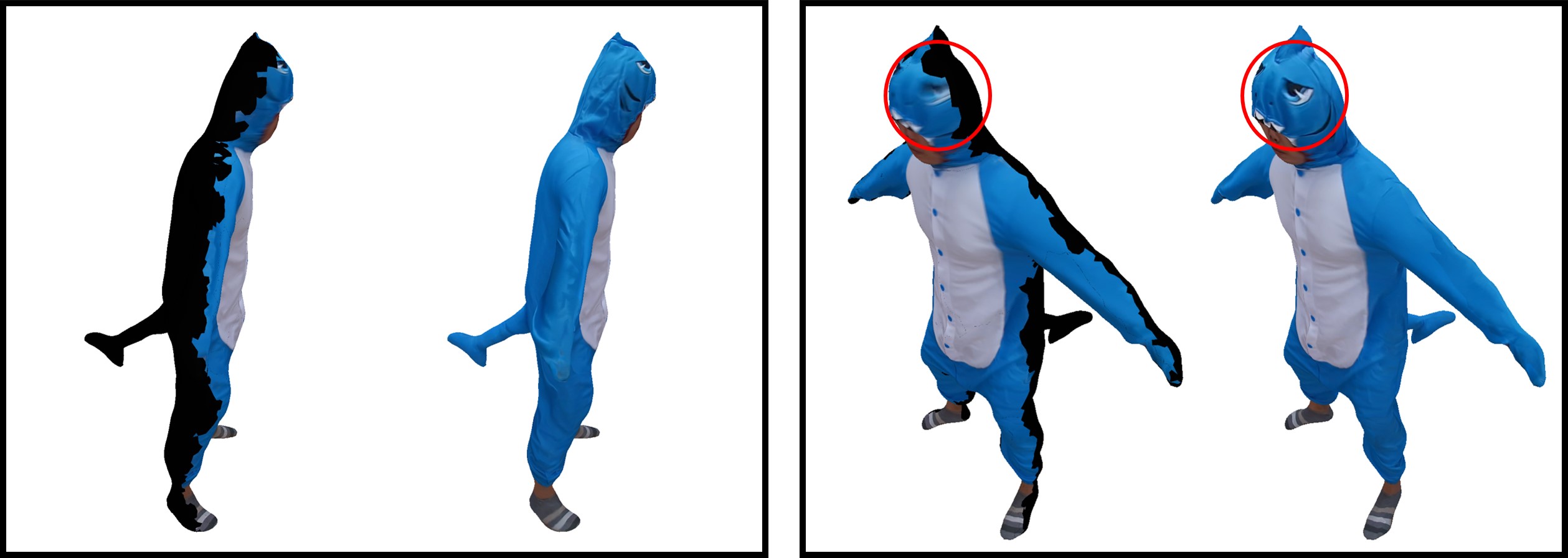} &
		\includegraphics[width=0.48\textwidth]{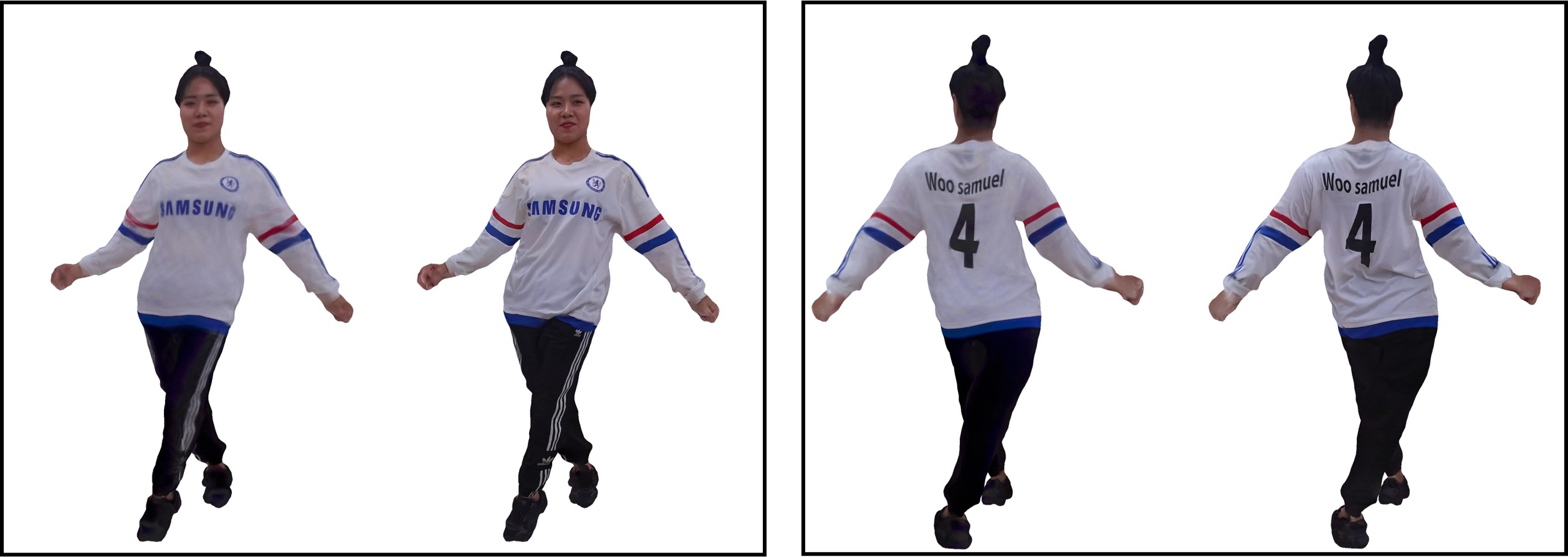} \\
		(a) comparison with single-view reconstruction & (b) comparison with global texture map~\cite{kim2019global} \\
	\end{tabular}
	\caption{Qualitative comparison. (a) (left) When single views are used for dynamic texture mapping, there are blank regions since the object is not fully observable at a single view. (right) Our approach reconstructs textures on the whole area since it produces a spatiotemporal volume containing every frame. Besides, our method produces sharp textures, even if there is a blurred part in an input image. (b) rendered images of a dynamic model using global texture map~\cite{kim2019global} (left) and our spatiotemporal texture (right). Our spatiotemporal texture reproduces wrinkles with reasonable directions.}
	\label{fig:comparison[qualitative comparison]}
\end{figure*}

\begin{figure*}[!ht]
	\centering
	\begin{tabular}{l@{\hspace{0.1cm}}c@{\hspace{0.2cm}}c}
		\includegraphics[width=0.2\textwidth]{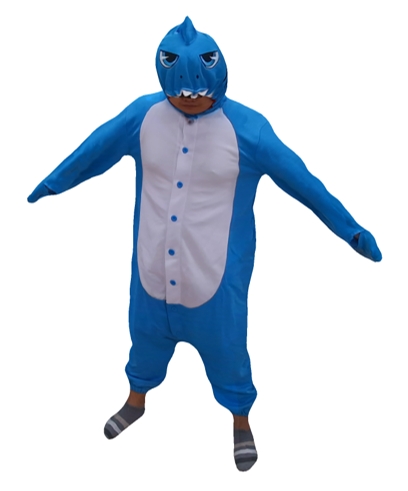}&
        \includegraphics[width=0.31\textwidth]{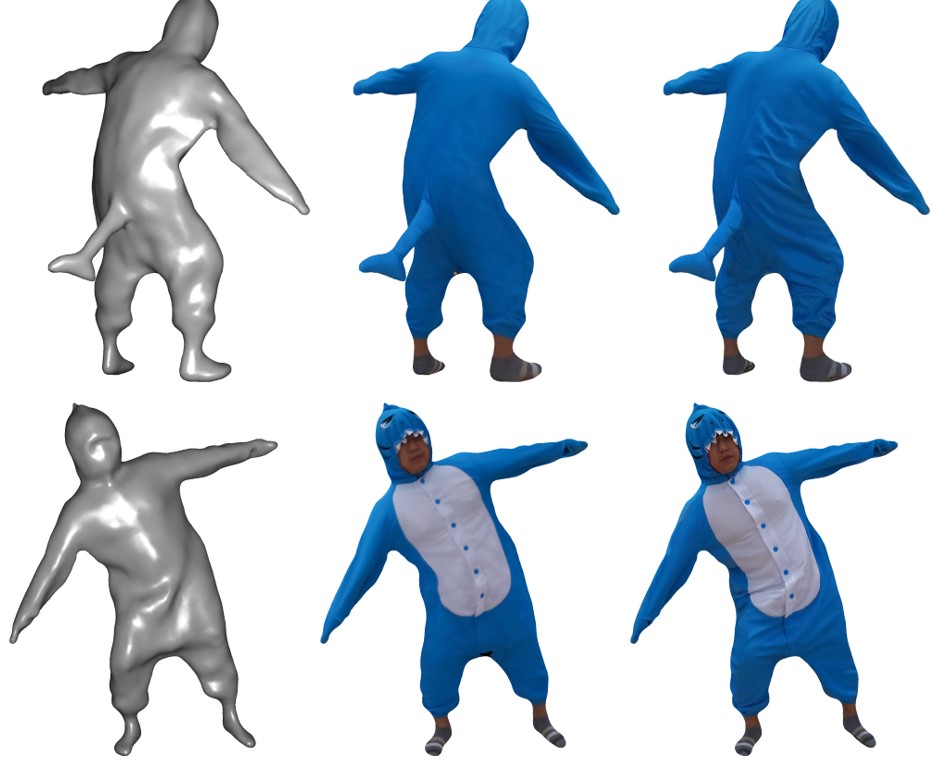} &
		\includegraphics[width=0.31\textwidth]{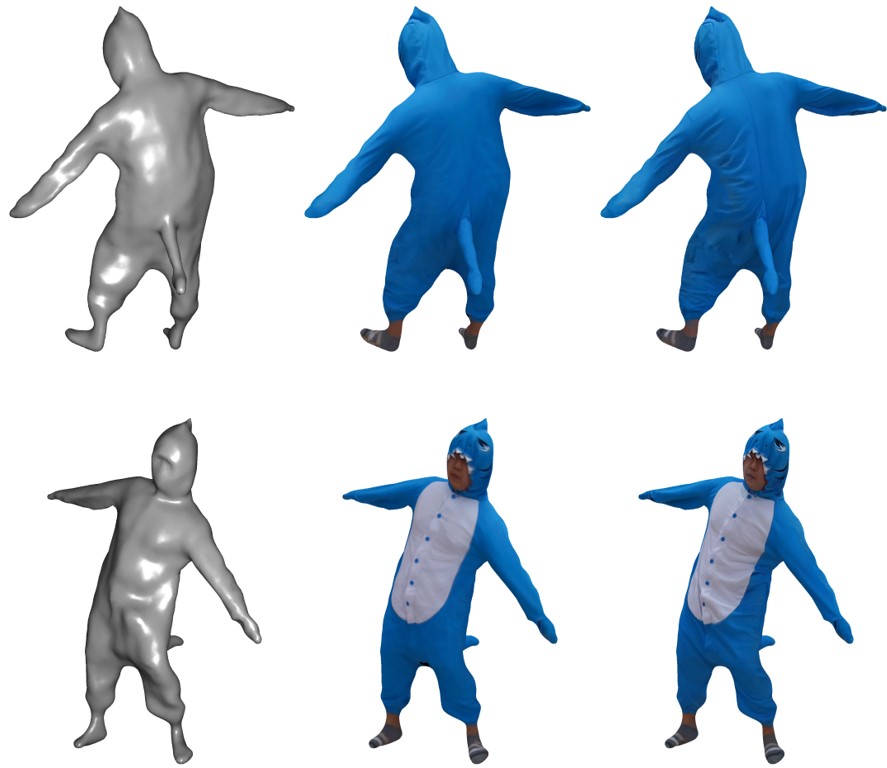}\\
		\includegraphics[width=0.165\textwidth]{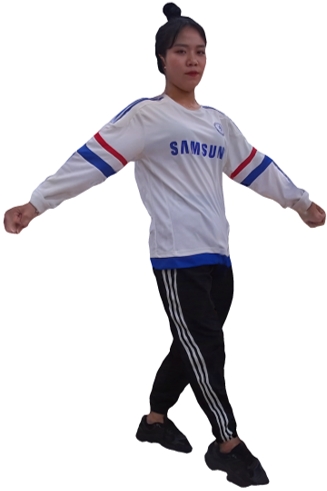}&
        \includegraphics[width=0.31\textwidth]{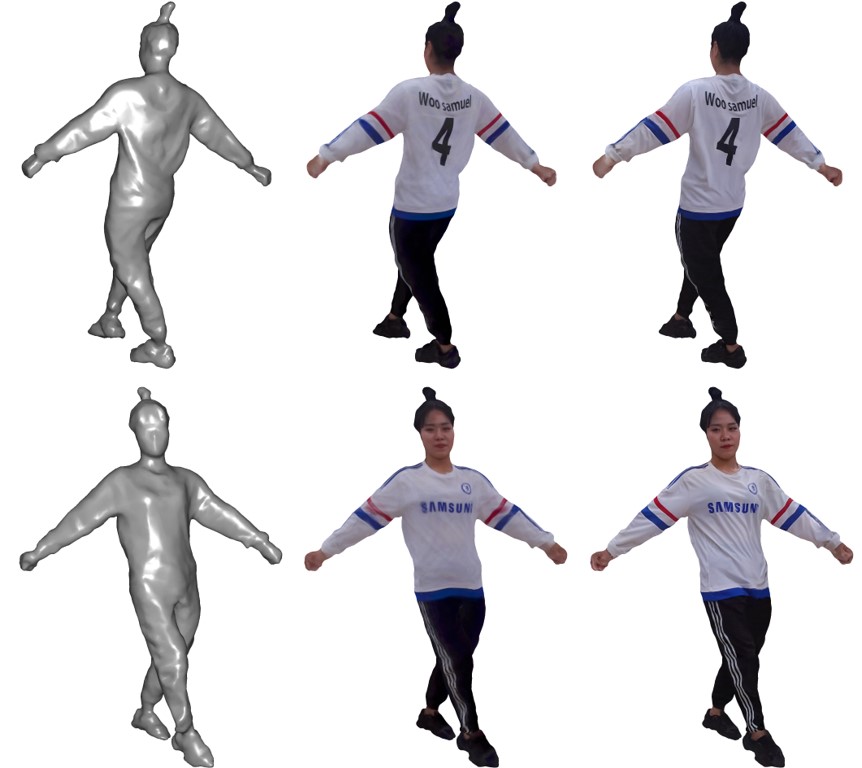} &
		\includegraphics[width=0.27\textwidth]{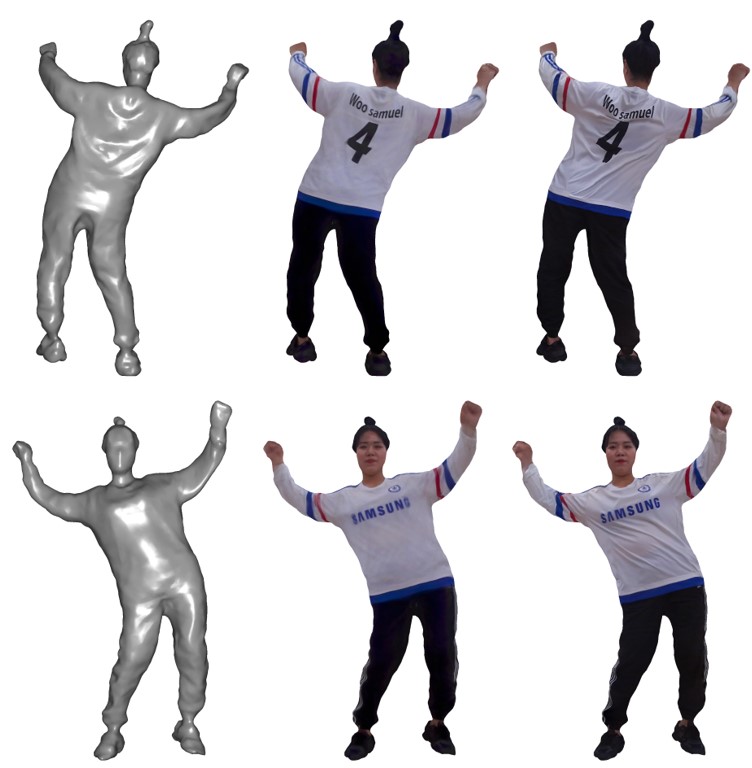}\\
	\end{tabular}
	\caption{Additional results. (left) snapshots of texture mapped geometry. (right) rear and front views of dynamic geometries at two timestamps. Each image trio shows the underlying geometry, global texture map-based result, and our result.}
	\label{fig:results[views]}
\end{figure*}

In \Fig{comparison[qualitative comparison]}a, we compare our results with a texture-mapped model using single images. The single image-based texture has black regions since some parts are invisible at a single frame, while our texture map covers the whole surface of the object.
Compared to global texture map~\cite{kim2019global}, our spatiotemporal texture is more appropriate to convey temporal contents (\Fig{comparison[qualitative comparison]}b). In particular, wrinkles on the clothes are expressed better. Additional results are shown in \Figs{results} and \ref{fig:results[views]}.

\subsection{User Study}

For a spatiotemporal texture reconstructed using our method, texture quality of originally visible regions can be measured by comparing with the corresponding regions in the input images, as we did in \Sec{quantitative}.
However, for regions invisible in the input images, there are no clear ground truths.
Even if an additional camera is set up to capture those invisible regions, the captured textures are not necessarily the only solutions, as various textures can be applied to a single shape, e.g., eyes may be open or closed with the same head posture.

\begin{figure}[t]
    \vspace*{-15pt}
	\centering
	\begin{tabular}{c@{\hspace{0.1cm}}c@{\hspace{0.1cm}}c}
		\includegraphics[width=0.23\textwidth]{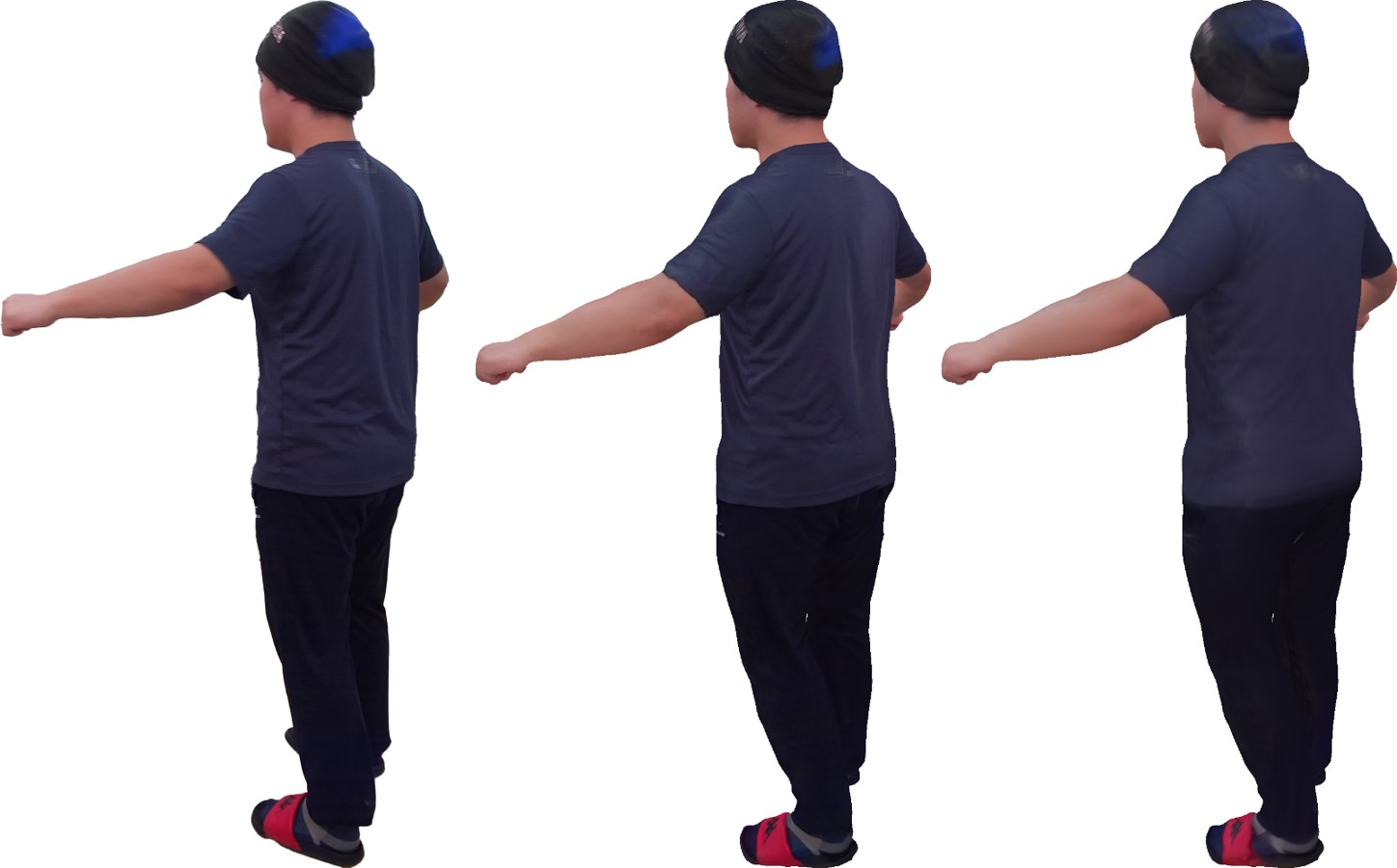}&
	    \includegraphics[width=0.22\textwidth]{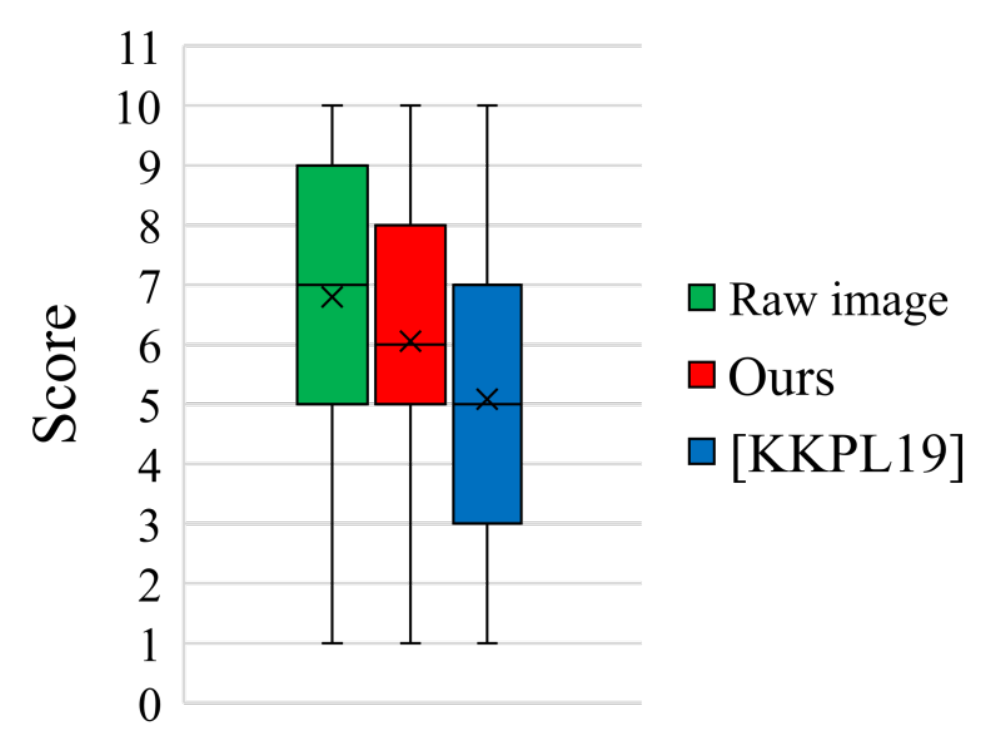}\\
		(a) sample image triple & (b) scoring result summary
	\end{tabular}
	\caption{User study. The image triple in (a) consists of an input image with background removed, a rendered dynamic mesh using spatiotemporal texture, a rendered dynamic mesh using global texture~\cite{kim2019global}, from left to right. In (b), from left to right, the averages are 6.798, 6.053, and 5.084, and the standard deviations are 2.4197, 2.3138, and 2.3653.}
	\label{fig:User_score_sample}
\end{figure}

To evaluate the quality of reconstructed textures for originally invisible regions without ground truths, we conducted a user study using Amazon Mechanical Turk (AMT).
We generated a set of triple images using input images, dynamic model rendered with spatiotemporal texture, and dynamic model rendered with global texture~\cite{kim2019global}.
We picked the triple images to be as similar as possible to each other, while removing the backgrounds of input images. \Fig{User_score_sample}a shows an example.
We prepared 40 image triples (120 images) and asked to evaluate whether the given images are realistic or not in two ways: One task is to choose a more realistic image among a pair of images. The other is to score how realistic a given single image is on a scale of 1-10. We set 25 workers whose hit approval rates are over 98\% to answer a single question. 
As a result, we collected total 3,000 pairwise comparison results and 3,000 scoring results. 

The preference matrix in \Tbl{preferencemat} summarizing the pairwise comparison results shows that the input images are clearly preferred to rendered images using reconstructed textures. This result would be inevitable as geometric reconstruction errors have introduced unnaturalness on top of texture reconstruction artifacts.
The errors are mainly caused by the template deformation method, which may not completely reconstruct geometric details. For example, in \Fig{User_score_sample}a, the sleeve of T-shirt is tightly attached to the arm in the reconstructed mesh.
Still, \Tbl{preferencemat} shows our spatiotemporal textures dominate global textures, where only texture reconstruction qualities are compared.
The scoring results summarized in \Fig{User_score_sample}b show similar conclusions.

\begin{table}[t]
    \centering
    \begin{tabular}{ c | c c c }
        \hline
        & Raw image & ~\cite{kim2019global} & Ours \\
        \hline
        Raw image & - & 863 & 718\\
        ~\cite{kim2019global} & 137 & - & 190\\
        Ours & 182 & 810 & -\\
        \hline 
    \end{tabular}
    \caption{Preference matrix from the user study. The value of each row refers to the number of times the image on that row is preferred when compared with the image on a particular column. 
    }
    \vspace*{-12pt}
    \label{tbl:preferencemat}
\end{table}
\begin{figure}[t]
	\vspace*{-15pt}
	\centering
	\begin{tabular}{c@{\hspace{0.1cm}}c@{\hspace{0.2cm}}c}
		\includegraphics[width=0.086\textwidth]{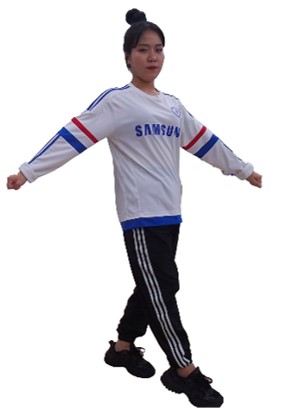} &
		\includegraphics[width=0.166\textwidth]{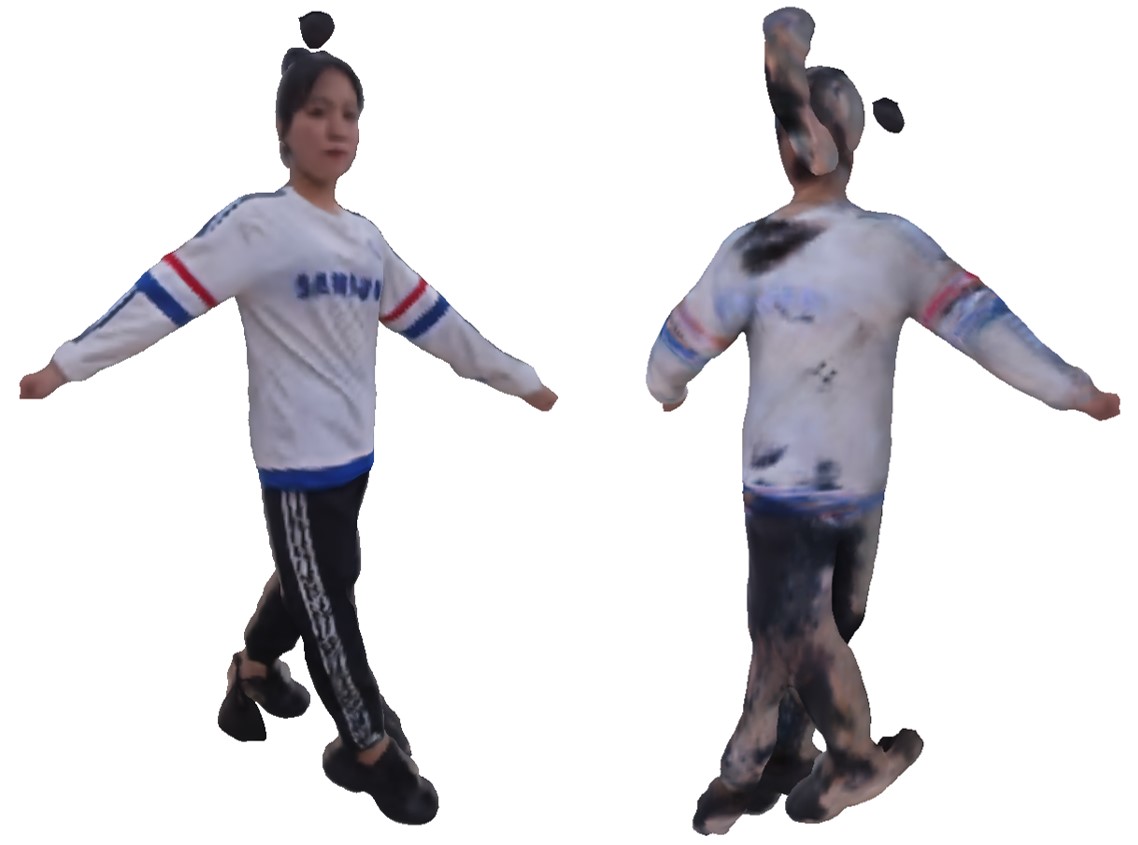} &
		\includegraphics[width=0.166\textwidth]{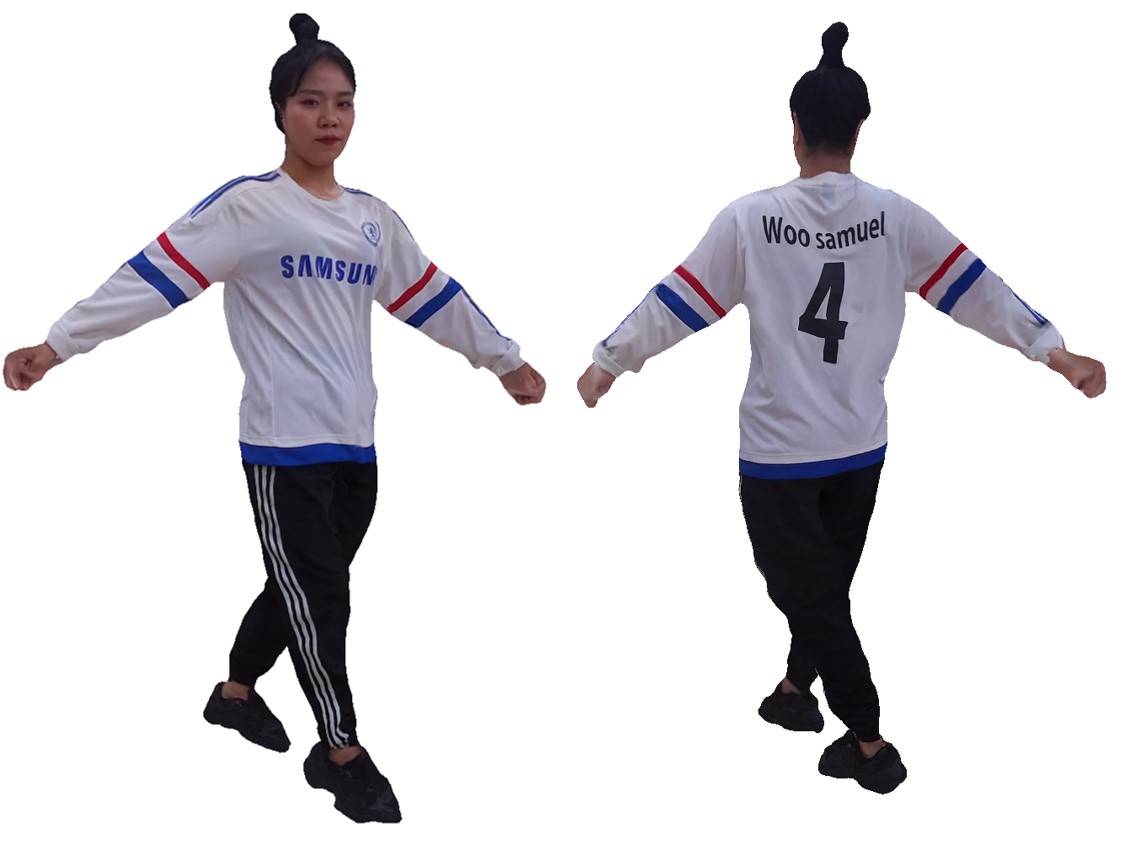}\\
		(a) Target frame & (b) PIFu~\cite{saito2019pifu} & (c) Ours\\
	\end{tabular}
	\caption{Comparison with a learning-based method. 
	Textures for the target frame (a) are reconstructed by PIFu (b) and our method (c).
    In PIFu, the input is a single RGB image with low resolution. The reconstructed textures are blurry (b left) for the visible regions in the input, and contain artifacts (b right) for the invisible regions. 
    }
	\label{fig:comparison[learning]}
	\vspace*{-13pt}
\end{figure}

\subsection{Comparison with Learning-based Methods}
\begin{figure*}[!t]
	\centering
	\begin{tabular}{c}
		\includegraphics[width=0.85\textwidth]{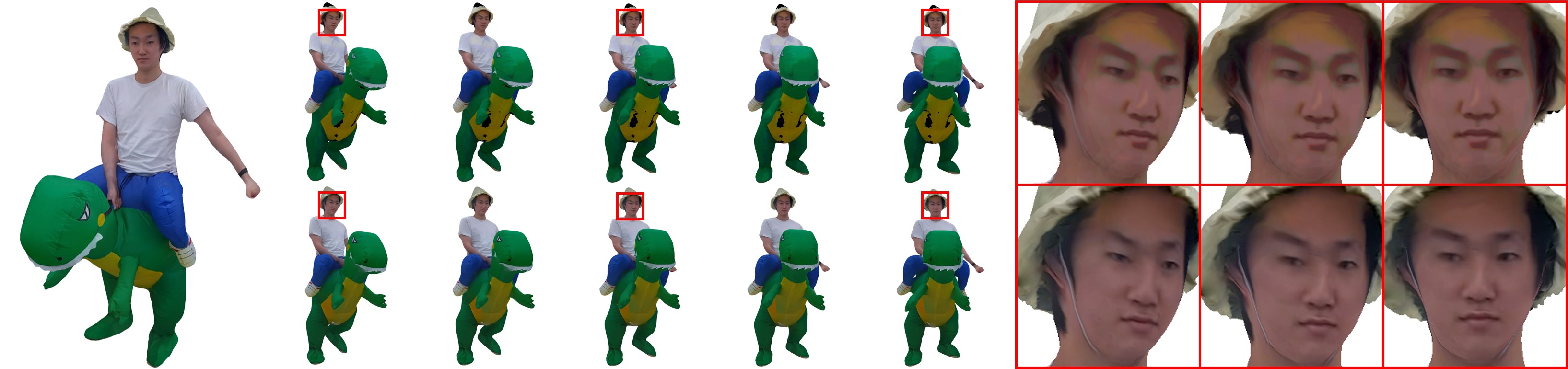}\\
		\includegraphics[width=0.85\textwidth]{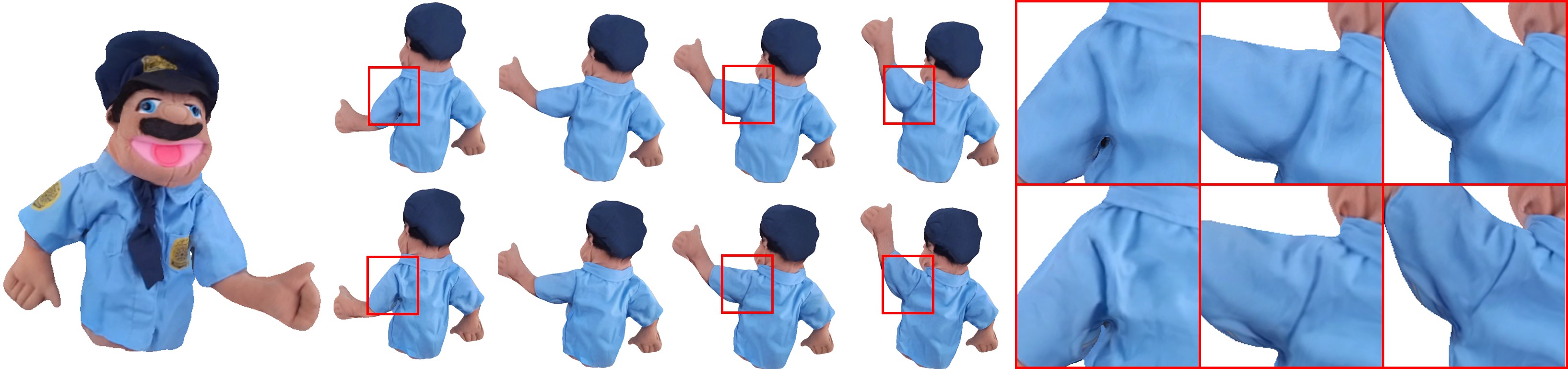}\\
  		\includegraphics[width=0.85\textwidth]{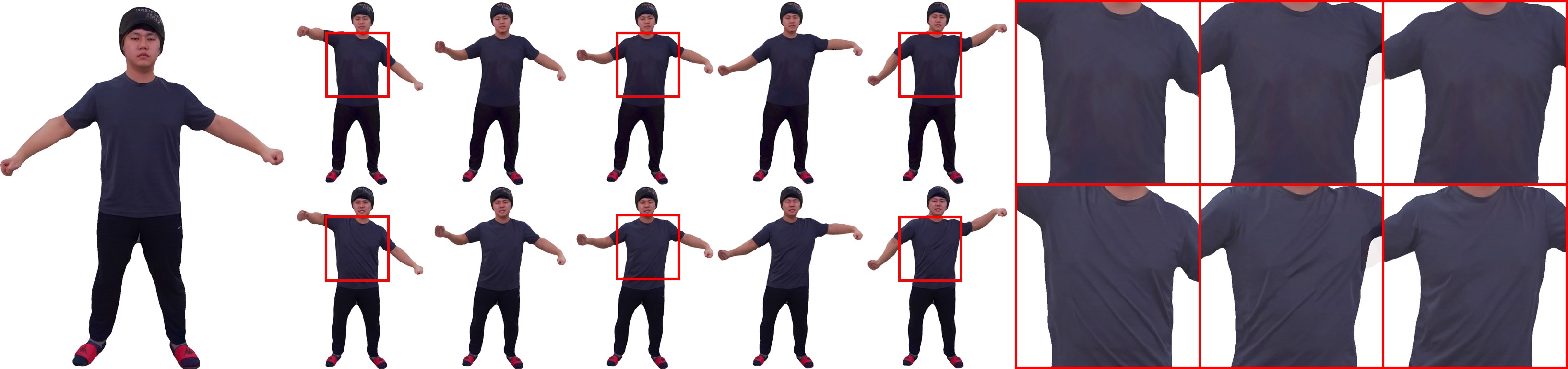}\\
		\includegraphics[width=0.85\textwidth]{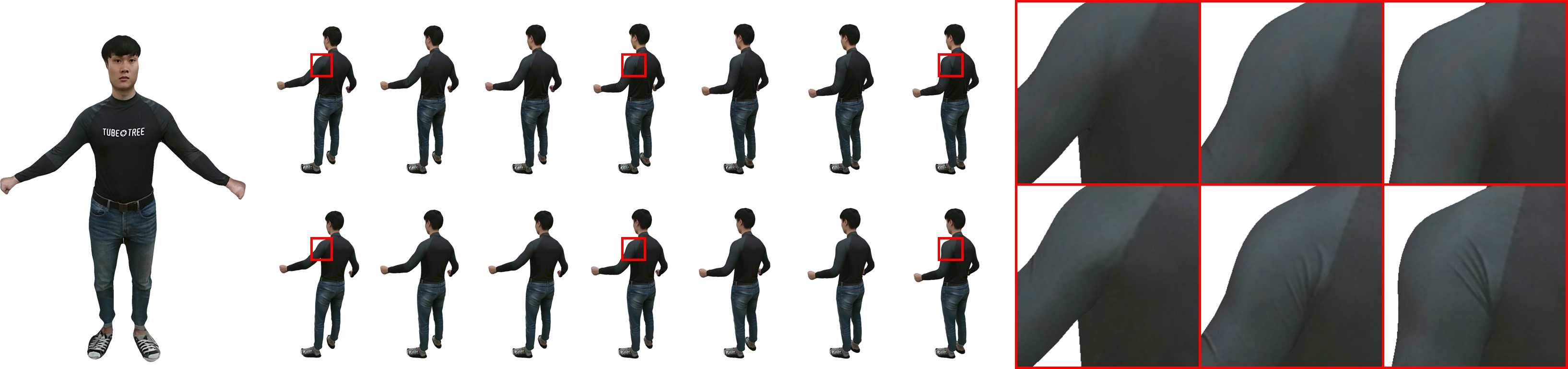}\\
	\end{tabular}
	\caption{Additional results. For each example, the top row shows the results from the global texture mapping based approach. The bottom row shows our results. Note that every rendered image was captured from the opposite side of the actual camera's position in the frame.}
	\label{fig:results}
\end{figure*}

As mentioned in \Sec{Related_Work}, there are learning-based methods to reconstruct textures of objects.
However, there are important differences between our method and those methods. Firstly, our method uses a template mesh and an image sequence from a single camera as the input, whereas the learning-based methods use a single image \cite{saito2019pifu, pandey2019volumetric} or multi-view images \cite{saito2019pifu}. Although it is unfair, we tested the authors' implementation of \cite{saito2019pifu} on single images of our dataset (\Fig{comparison[learning]}). The generated textures on the originally invisible areas are clearly worse than our results. Note that multi-view images that can be used in \cite{saito2019pifu} 
should be captured by multiple cameras at the same time. Then the setting differs from our case of a single camera, making comparison using multi-view images inappropriate. Secondly, due to memory limitation, the networks of \cite{saito2019pifu, pandey2019volumetric} cannot fully utilize the textures of our 4K dataset. This would lead to poor quality compared to our results. Thirdly, the learning-based methods target only humans, while our method does not have the limitation, as demonstrated in the top two examples of \Fig{results}.
Finally, our method reconstructs a completed texture volume that can be readily used for generating consistent novel views in a traditional rendering pipeline, but synthesized novel views from learning-based methods are not guaranteed to be consistent.

\begin{figure}[t]
	\centering
	\begin{tabular}{c@{\hspace{0.1cm}}c@{\hspace{0.1cm}}c@{\hspace{0.1cm}}c}
		\includegraphics[width=0.11\textwidth]{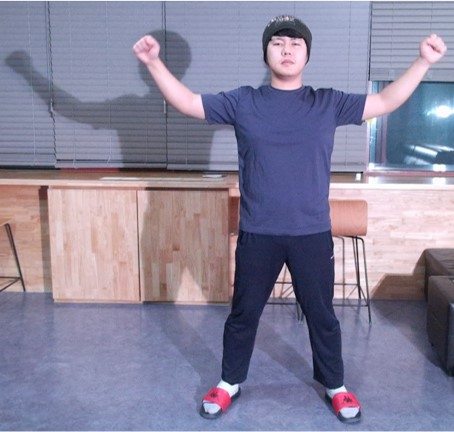} &
		\includegraphics[width=0.11\textwidth]{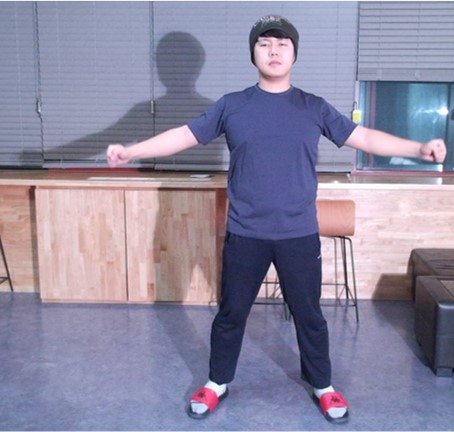} &
		\includegraphics[width=0.11\textwidth]{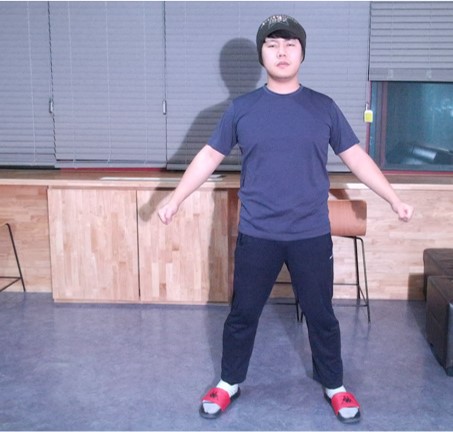} &
		\includegraphics[width=0.11\textwidth]{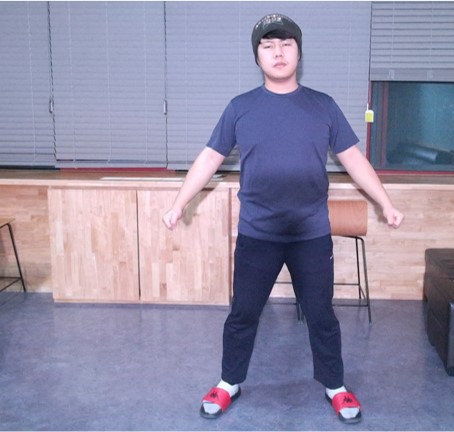}\\
		(a) frame 50 & (b) frame 60 & (c) frame 70 & (d) frame 80\\
	\end{tabular}
	\begin{tabular}{c}
		\includegraphics[width=0.45\textwidth]{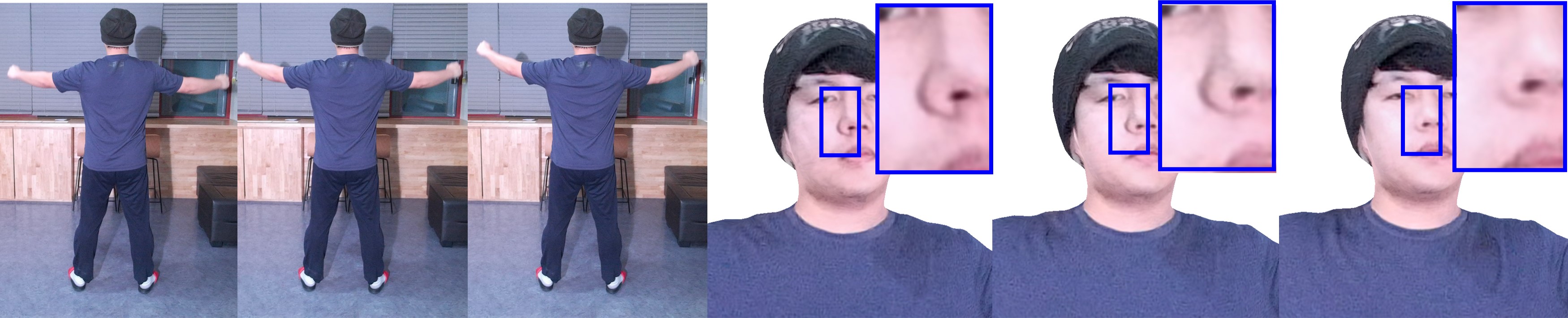}\\
		(e) input images and reconstructed textures on frames 160-162.\\
	\end{tabular}
	\caption{Variant lighting example. 
	A directional light moves continuously in the input images (see the movement of shadow).
	However, the reconstructed textures in (e) on invisible regions show sudden brightness changes.}
	\label{fig:Fail[varlight]}
	\vspace*{-14pt}
\end{figure}

\subsection{Discussions and Limitations}
\label{sec:7_limitations}
\begin{figure}[t]
	\centering
	\includegraphics[width=0.2\textwidth]{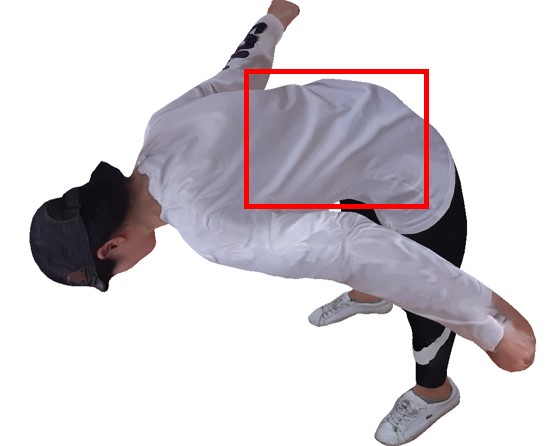}
	\caption{Failure case of an unseen region. The red boxed region is never visible from any input color image, and the recovered texture is unnatural.
	}
	\label{fig:Limitation}
	\vspace*{-8pt}
\end{figure}

\paragraph*{Rich textures}
Our method would work for rich textures as far as our texture coordinate optimization works properly. Texture coordinate optimization may fail in an extreme case. For example, when the texture is densely repeated, the misalignment in geometric reconstruction could be larger than the texture repetition period. Then, the texture correspondence among different frames could be wrongly optimized.
\paragraph*{Computation time}
Our approach takes quite some time and it would be hard to become real-time. On the other hand, texture reconstruction for a model is needed to be done only once, and our method produces high-quality results that can be readily used in a conventional rendering pipeline.
\paragraph*{Dependency on geometry processing}
If non-rigid registration of dynamic geometry fails completely, our texture reconstruction fails accordingly. However, the main focus of this paper is to reconstruct high-quality textures, and our method would be benefited from the advances of single-view non-rigid registration techniques. In our approach, texture coordinate optimization helps handle possible misalignments from non-rigid registration of various motions. For fast motions, as shown in \Fig{quan[blur]}, our labeling data term could filter out blurry textures, replacing them with sharper ones.
\paragraph*{Fixed topology}
As template-based non-rigid models have been steadily researched until recently~\cite{li2018robust, yao2020quasi}, our method assumes that each model in the motion sequence has the same fixed topology. As a result, our method may not handle extreme shape changes, such as taking off clothes or looser clothes.
\paragraph*{Variant lights and relighting}
Our method may not produce plausible texture brightness changes in a variant lighting environment.
In \Fig{Fail[varlight]}, a directional light is continuously moving when the input images are captured. However, the reconstructed textures on invisible regions show sudden brightness changes at some frames. As a result, sometimes temporal flickering can be found on originally invisible regions. 
Another limitation of our method is that lighting is not separated from textures but included in the restored textures. This limitation would hinder relighting of the textured objects, but could be resolved by performing intrinsic decomposition on the input images.
\paragraph*{Unseen textures}
Our method cannot generate textures that are not contained in the input images. 
Unnatural textures could be produced for regions invisible at all frames (\Fig{Limitation}).

\section{Conclusions}  
\label{sec:8_conclusion}
We have presented a novel framework for generating a time-varying texture map (spatiotemporal texture) for a dynamic 3D model using a single RGB-D camera. Our approach adjusts the texture coordinates and selects an effective frame label for each face of the model. The proposed framework generates plausible textures  conforming to shape changes of a dynamic model.

There are several ways to improve our approach. First, to conduct the labeling process more efficiently, considering a specific period like an approach by Liao~\cite{liao2013automated} would be a viable option. Second, our framework separates the 3D registration part and texture map generation part. We expect better results if the geometry and texture are optimized jointly~\cite{fu2020joint}.

\vspace{-12px}
\subsection*{Acknowledgements}
This work was supported by the Ministry of Science and ICT, Korea, through IITP grant (IITP-2015-0-00174), NRF grant (NRF-2017M3C4A7066317), and Artificial Intelligence Graduate School Program (POSTECH, 2019-0-01906).


\newcommand{\etalchar}[1]{$^{#1}$}


\end{document}